\documentclass[10pt,twocolumn,letterpaper]{article}

\usepackage{cvpr}              

\usepackage{multirow}
\usepackage{graphicx}
\usepackage{booktabs}
\usepackage{wrapfig}
\usepackage{subcaption}
\usepackage{bm}
\usepackage{academicons}
\usepackage{fontawesome}
\usepackage{placeins}

\usepackage[table]{xcolor}
\definecolor{lighttan}{RGB}{255, 235, 210}
\definecolor{lightgreen}{HTML}{C8E6C9}   
\definecolor{lightblue}{HTML}{BBDEFB}    
\definecolor{lightyellow}{HTML}{FFF9C4}  
\definecolor{lightorange}{HTML}{FFE0B2}  
\definecolor{lightred}{HTML}{FFCDD2}     
\definecolor{lightpurple}{HTML}{E1BEE7}  
\definecolor{lightgray}{HTML}{F5F5F5}    
\definecolor{lightteal}{HTML}{B2DFDB}    
\definecolor{lightsage}{HTML}{DCECC9}    

\usepackage[most]{tcolorbox}
\definecolor{light_green}{HTML}{b2ffb2}
\definecolor{sage}{HTML}{c3efb2}
\definecolor{light_red}{HTML}{ffb2b2}
\definecolor{light_blue}{HTML}{add8e6}
\tcbset{on line, 
        boxsep=1pt, left=1pt,right=1pt,top=1pt,bottom=1pt,
        colframe=white,
        colback=light_green,  
        highlight math style={enhanced}
        }

\DeclareUnicodeCharacter{2265}{$\ge$}
\DeclareUnicodeCharacter{2264}{$\le$} 

\definecolor{cvprblue}{rgb}{0.21,0.49,0.74}
\usepackage[pagebackref,breaklinks,colorlinks,allcolors=cvprblue]{hyperref}

\newcommand{\model}{MOMO}

\def\paperID{21599} 
\def\confName{CVPR}
\def\confYear{2026}

\title{
  \vspace{-1.0em}
  \begin{center}
    \begin{minipage}[c]{0.0001\textwidth}
      \centering
      \vspace{-0.5em}
      \includegraphics[height=4em]{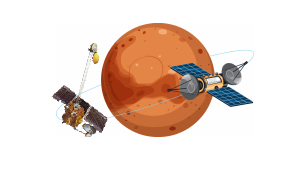}
    \end{minipage}%
    \hspace{2em}%
    \begin{minipage}[c]{0.85\textwidth}
      \centering 
      \textbf{\textcolor{purple}{\model}: \textcolor{purple}{M}ars \textcolor{purple}{O}rbital \textcolor{purple}{Mo}del\\
      Foundation Model for Mars Orbital Applications}
    \end{minipage}
  \end{center}
  \vspace{-1em}
}

\author{\textbf{Mirali Purohit}$^{1, 2}$\textsuperscript{ \faEnvelope} \quad \textbf{Bimal Gajera}$^{1*}$ \quad \textbf{Irish Mehta}$^{1*}$ \quad \textbf{Bhanu Tokas}$^{1*}$ \\ \textbf{Jacob Adler}$^1$ \quad \textbf{Steven Lu}$^2$ \quad \textbf{Scott Dickenshied}$^1$ \quad \textbf{Serina Diniega}$^2$ \\ \textbf{Brian Bue}$^2$ \quad \textbf{Umaa Rebbapragada}$^2$ \quad \textbf{Hannah Kerner}$^1$ \\
\vspace{0.4mm}\\
$^1$Arizona State University \\ $^2$Jet Propulsion Laboratory, California Institute of Technology
}

\begin{document}

\maketitle

\begin{abstract}

We introduce \model, the first multi-sensor foundation model for Mars remote sensing. \model{} uses model merge to integrate representations learned independently from three key Martian sensors (HiRISE, CTX, and THEMIS), spanning resolutions from 0.25 m/pixel to 100 m/pixel. Central to our method is our novel Equal Validation Loss (EVL) strategy, which aligns checkpoints across sensors based on validation loss similarity before fusion via task arithmetic. This ensures models are merged at compatible convergence stages, leading to improved stability and generalization. We train \model{} on a large-scale, high-quality corpus of $\sim12$ million samples curated from Mars orbital data and evaluate it on 9 downstream tasks from Mars-Bench. \model{} achieves better overall performance compared to ImageNet pre-trained, earth observation foundation model, sensor-specific pre-training, and fully-supervised baselines. Particularly on segmentation tasks, MOMO shows consistent and significant performance improvement. Our results demonstrate that model merging through an optimal checkpoint selection strategy provides an effective approach for building foundation models for multi-resolution data. The model weights, pretraining code, pretraining data, and evaluation code are available at: \href{https://github.com/kerner-lab/MOMO}{github.com/kerner-lab/MOMO}.

\def\thefootnote{\faEnvelope}\footnotetext{ Corresponding Author: mpurohi3@asu.edu}\def\thefootnote{\english{footnote}}
\def\thefootnote{*}\footnotetext{ Equal Contribution}\def\thefootnote{\english{footnote}}

\end{abstract}

\section{Introduction}
\label{sec:introduction}

Foundation models (FMs) have demonstrated strong capability in learning representations from large-scale data, enabling improved downstream task performance compared to conventional supervised training from scratch \cite{lu2025vision, huang2025survey}. In recent years, more than 150 FMs have been proposed for Earth observation (EO) applications \cite{huang2025survey, lu2025vision}. These EO-FMs are being actively used in applications such as food security, disaster response, and climate change \cite{rolnick2024application, butsko2025deploying}.

Similar to Earth-orbiting satellites, Mars-orbiting satellites systematically collect remote sensing observations of the planet's surface and atmosphere. In contrast to the active research on EO-FMs in recent years, no FM has been proposed for Mars remote sensing applications to date. An FM for Mars remote sensing would enable planetary scientists to train models for custom science tasks at a lower cost compared to fully supervised methods. Researchers are already using ImageNet pre-trained models for Mars remote sensing applications \cite{purohit2024conequest, wagstaff2021mars, wagstaff2018deep}, but in-domain pre-training could improve performance and generalization, as suggested by preliminary findings in \cite {purohit2024investigating}.

\begin{figure*}
    \centering
    \begin{minipage}[c]{0.68\linewidth}
        \includegraphics[width=\linewidth]{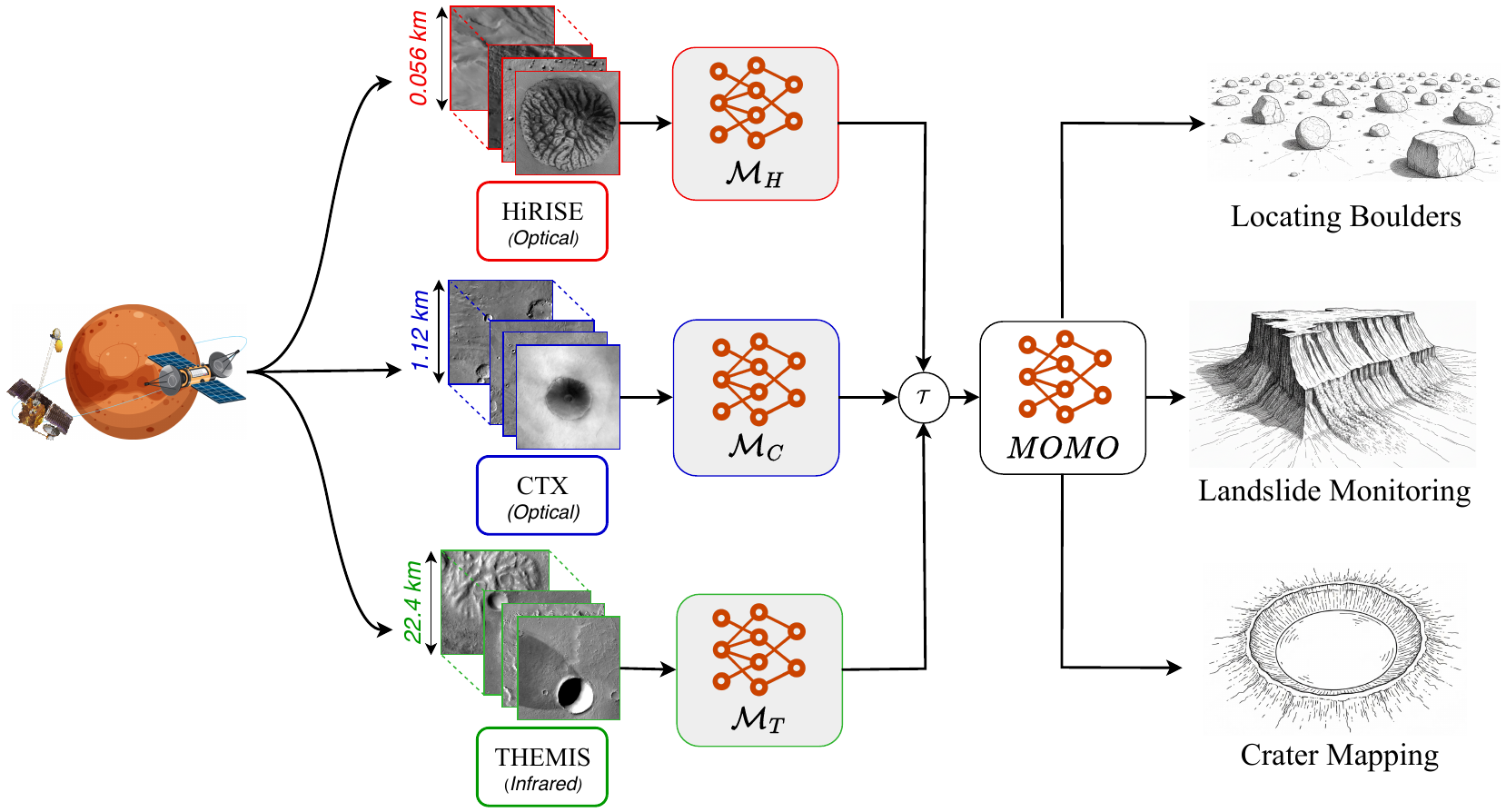}
    \end{minipage}
    \hfill
    \begin{minipage}[c]{0.3\linewidth}
        \caption{\model{} can be effectively applied across a wide range of resolutions and a broad spectrum of Martian remote sensing tasks. By leveraging diverse sensors, our approach enables a single model to generalize across different orbital applications, including large-scale crater or landslide mapping and precise boulder localization.}
        \label{fig:teaser_figure}
    \end{minipage}
\end{figure*}

Developing FMs for remote sensing data requires significant domain expertise and computational resources \cite{rolf2024position}. Satellite images are acquired from multiple sensors, each operating at different wavelengths, spatial resolutions, and spectral channels. The EO community has proposed customized model architectures specifically designed for Earth satellite data and applications \cite{tseng2023lightweight, astruc2024anysat, han2024bridging}. While effective for EO data, these approaches do not extend directly to Mars remote sensing data due to differences in sensor properties and availability.

We propose \textbf{\textcolor{purple}\model} (\textbf{\textcolor{purple}M}ars \textbf{\textcolor{purple}O}rbital \textbf{\textcolor{purple}{Mo}}del), the first FM for Mars remote sensing applications. We introduce a novel approach to handle data efficiently from multiple sensors that measure different physical properties at different spatial scales. We pre-train individual models on data from each sensor and subsequently merge them using our novel checkpoint selection strategy. We evaluate \model{}  on all 9 orbital downstream tasks from Mars-Bench \cite{purohit2025marsbench}. We compare our proposed method to a range of strong baselines. Overall, \model{} outperforms models pre-trained on Earth Observation data, sensor-specific Mars data, and ImageNet, demonstrating superior performance on segmentation tasks and comparable performance on classification. In summary, our main contributions are as follows:
\begin{itemize}
    \item We introduce \model, the first foundation model for Mars orbital applications. \model{} efficiently handles multi-sensor and multi-resolution data. To the best of our knowledge, this is the \textit{first} systematically developed, analyzed, and evaluated foundation model for Mars tasks.
    \item We propose a novel technique to build a multi-sensor foundation model through model merging, using our optimal checkpoint selection strategy for stable fusion.
    \item We conduct extensive comparisons of \model{} against multiple baselines, including ImageNet pre-training, Earth-observation foundation models, sensor-specific pre-training, and other checkpoint selection strategies. Our results on Mars-Bench demonstrate that \model{} achieves superior overall performance across tasks.
\end{itemize}


\section{Related Work}
\label{sec:related_work}

\paragraph{Self-Supervised Learning for Mars Orbital Tasks.} There are a few preliminary studies that have explored self-supervised learning for Mars remote sensing. Jiang et al. pre-trained a model on just 13 HiRISE images for landform detection \cite{jiang2021robust}, but their focus is only on the landmark detection task. In contrast, Purohit et al. pre-trained a model on 1 million CTX image patches and demonstrated that in-domain pre-training can surpass ImageNet pre-training for certain tasks \cite{purohit2024investigating}. However, this work only uses data from a single instrument and evaluates on 2 downstream tasks.

\paragraph{Foundation Models for Earth Observation.} Over the past 4-5 years, researchers have introduced numerous FMs for EO applications. Many of these include masked autoencoder-based models such as SatMAE \cite{cong2022satmae}, ScaleMAE \cite{reed2023scale}, SatMAE++ \cite{Noman_2024_CVPR}, Presto \cite{tseng2023lightweight}; contrastive learning-based models like SeCo \cite{manas2021seasonal}, SatCLIP \cite{klemmer2023satclip}, GeoCLIP \cite{vivanco2023geoclip}, CROMA \cite{fuller2024croma}; and other approaches including msGFM \cite{han2024bridging}, AnySat \cite{astruc2024anysat}, Galileo \cite{tseng2025galileo}, Satlas \cite{bastani2023satlaspretrain}, SkySense \cite{Guo_2024_CVPR}, SpectralGPT \cite{hong2024spectralgpt}, and SpectralEarth \cite{braham2024spectralearth}. Earth and Mars remote sensing data have some similarities, such as multispectral/multi-sensor observations and overhead imaging geometry, but they are very different. Mars has different atmospheric conditions, illumination, surface materials, and sensor characteristics. We would not expect EO-FMs to generalize to Mars remote sensing tasks.

\vspace{-2mm}

\paragraph{Model-editing.} Model-editing methods aim to enhance generalization, robustness, and out-of-distribution performance without incurring extra inference costs. Model editing has been applied across various domains and tasks. Techniques include merging weights of multiple fine-tuned models via simple weight averaging or ensembling, layer-wise matching, and soft alignment; to create a single model that outperforms its individual components. There are dozens of proposed research methods based on model averaging \cite{szegedy2015going, izmailov2018averaging, foret2020sharpness, wortsman2022model, li2022trainable, guo2023stochastic, kaddour2022stop, li2022branch} and ensembling techniques, combining the outputs of two or more models \cite{bauer1999empirical, lakshminarayanan2017simple, freund1997decision, ovadia2019can, mendoza2016towards, wenzel2020hyperparameter}. Researchers have studied more efficient methods for modifying a model’s behavior through interventions after pre-training, referring to this process by different names, such as patching \cite{goel2020model, ilharco2022patching, murty2022fixing, sung2021training}, editing \cite{santurkar2021editing, mitchell2021fast, mitchell2022memory}, aligning \cite{askell2021general, ouyang2022training}, and layer-wise editing \cite{stoica2023zipit}.

Prior work has focused on merging models trained on similar or different distributions, but limited focus has been given to checkpoint selection before merging. Existing methods typically merge models at their final checkpoints without considering differences in training trajectories. 

In contrast, we introduce a novel checkpoint selection strategy based on validation loss alignment, which ensures models trained on different data distributions (in our case, varying spatial resolutions) are merged at their most compatible stage. To the best of our knowledge, this is the first task-arithmetic approach to introduce a systematic checkpoint selection strategy.

\section{\model}
\label{sec:MOMO}

\subsection{Motivation}
\label{subsec:motivation}

Before describing our methodology, we first provide a brief overview of how foundation models are typically developed in Earth Observation and how our approach differs. Most EO-FMs take one of two approaches to combine data from multiple sensors: 1) stacking spatially- and temporally-aligned data from each sensor as different channels to a single encoder or a separate tokenizer for each sensor (e.g., \cite{tseng2025galileo, cong2022satmae, tseng2023lightweight}), or 2) combining data from multiple sensors in a single heterogeneous pre-training dataset (e.g., \cite{bastani2023satlaspretrain}). 

The stacking approach requires a large number of coincident (spatially and temporally overlapping) observations from each sensor and sensors with somewhat similar spatial resolutions (e.g., 10-30 m/pixel). Mars orbital sensors have very different coverage (e.g., CTX covers nearly 100\% of the planet~\cite{dickson2023release}, but HiRISE only covers less than 3\%~\cite{mcewen2024high}) and spatial resolutions (e.g., THEMIS is 100 m/pixel compared to HiRISE at 0.25 m/pixel; see Figure~\ref{fig:teaser_figure}). The data-combination approach is feasible, but would require training a new model whenever a new sensor is added. 

To address these challenges, we propose a methodology that avoids directly combining heterogeneous data from multiple sensors. Instead, we pre-train independent masked autoencoder models for each sensor, allowing each model to first learn the unique distribution and feature characteristics of its respective sensor. We then merge these sensor-specific models (Figure \ref{fig:teaser_figure}) using our proposed Equal Validation Loss (EVL) strategy, which aligns checkpoints based on validation loss similarity to ensure compatibility before fusion. This is the first work to employ a model-merging strategy to construct a remote sensing foundation model. Additionally, we introduce a customized cost function designed to optimize the reconstruction process by capturing both pixel-level and perceptual information. Full methodological details are provided in the following sections.

\subsection{Cost Function}
\label{subsec:cost_function}

Loss functions play a key role in training deep learning models. An unsuitable objective function can lead to convergence toward suboptimal local minima or undesirable optimization directions. 

Following widely adopted practices for training Masked AutoEncoders (MAEs) \cite{he2022masked}, we initially pre-trained our model using mean squared error (MSE) as the reconstruction objective. However, after visualizing the reconstructed outputs, we observed that while the model could accurately recover the color distribution and surface textures of masked regions, it often failed to reconstruct structural details of key geomorphologic features (e.g., the accurate shape of a crater) when such regions were masked (sample reconstructions in Appendix \ref{subsec:reconstruction}). 

This limitation arises because MSE is a pixel-level loss, emphasizing low-level intensity matching rather than perceptual or structural fidelity. Although MSE effectively captures color and tone consistency, it lacks sensitivity to higher-order spatial features such as shape, boundary continuity, and object geometry.

To address this, we introduce additional perceptual and structure-aware components in our loss function to guide the model toward learning edge-level and shape-consistent representations. We add terms that minimize LPIPS \cite{zhang2018unreasonable} and maximize structural similarity (SSIM) \cite{nilsson2020understanding}. In addition, to enforce spatial smoothness and structural consistency, we penalize the difference between horizontal and vertical gradients of predicted and ground-truth images \cite{stoica2023zipit, ma2020structure}. For an image $I$ and its reconstruction $\hat{I}$, the gradient loss is formulated as an $\ell_1$ penalty:
\begin{equation}
    \mathcal{L}_{\text{grad}} = \frac{1}{N} \sum_{i,j} \Big( \left| \partial_x I_{i,j} - \partial_x \hat{I}_{i,j} \right| + \left| \partial_y I_{i,j} - \partial_y \hat{I}_{i,j} \right| \Big)
\end{equation}

The final combined pre-training objective function is:
\begin{equation}
    \mathcal{L}_{\text{total}} = \lambda_1 \mathcal{L}_{\text{MSE}} + \lambda_2 \mathcal{L}_{\text{SSIM}} + \lambda_3 \mathcal{L}_{\text{LPIPS}} + \lambda_4 \mathcal{L}_{\text{grad}},
\label{eq:overall_loss}
\end{equation}

\noindent where $\mathcal{L}_{\text{SSIM}}$ is $(1- SSIM)$ and $\lambda_i$ are weighting coefficients.
By combining pixel-wise, perceptual, and gradient-aware objectives, \model{} achieves reconstructions that preserve both spatial and structural details, particularly important for Martian features such as crater rims, cones, and landslide boundaries.

\subsection{Optimal Checkpoint Selection Strategy}
\label{subsec:checkpoint}

We define our checkpoint selection strategy as follows: Let there be $n$ distinct sensors, each associated with a dataset $\mathcal{D}_i$ $(i = 1, 2, \dots, n)$ representing distinct spatial resolutions, modalities, or imaging characteristics. We train $n$ independent models $\{\mathcal{M}_i\}_{i=1}^{n}$, where each $\mathcal{M}_i$ is optimized on $\mathcal{D}_i$ for $k$ epochs denoted as $E = \{e_1, e_2, e_3, ..., e_k\}$. During training, we record the validation loss $\mathcal{L}_i^{(e)}$ for every epoch $e \in E$. Here, each model is pre-trained by optimizing the loss, which is defined in Equation \ref{eq:overall_loss}.

Instead of merging all models at their final checkpoints, we introduce the \textbf{Equal Validation Loss (EVL)} strategy, which aligns checkpoints across sensors based on validation loss similarity prior to model fusion. This ensures that models trained on heterogeneous data distributions are combined at a mutually compatible convergence stage.

\vspace{-3mm}

\paragraph{Loss alignment.}
Let $\mathcal{L}_i^{(e)}$ denote the validation loss of sensor $i$ at epoch $e$,  for $i \in \{1, \dots, n\}$ and $e \in E$. We define a \emph{candidate epoch tuple},
\[
\mathbf{t_c} = (e^1, e^2, \dots, e^n),
\]
where $e^i$ denotes the selected epoch for model $\mathcal{M}_i$. To create a candidate tuple, we iterate over epoch indices and form candidate epoch tuples by taking one epoch from each sensor and checking whether the losses in that tuple are mutually close. A tuple $\mathbf{t_c}$ is considered \emph{loss-aligned} if the validation losses across all sensors are mutually close, i.e.,

\[
\Delta_{ij} =  \bigl|\,\mathcal{L}_i^{(e^{i}_{_a})} - \mathcal{L}_j^{(e^{j}_{b})}\bigr|
\]
\begin{equation*}
\label{eq:sensor-consistency}
\forall\, i,j \in \{1,\dots,n\};\ a,b \in \{1,\dots,k\}; i \neq j;\ \Delta_{ij} \le \epsilon
\end{equation*}
where $e^{i}_{a}$ represents the $a^{th}$ epoch on the model trained on $i^{th}$ sensor. $\epsilon > 0$ is a small tolerance hyperparameter. The set of all loss-aligned tuples $t_{c}$ is denoted as $\mathcal{E}_{\text{EVL}}$. 

\paragraph{Distance-guided checkpoint selection.}
For each loss-aligned tuple $\mathbf{t_c}\in\mathcal{E}_{\mathrm{EVL}}$, we measure how far the selected epochs deviate from their respective early-stopping epochs ($s_{es}$). We define the normalized average epoch distance as
\[
\bar{D}(\mathbf{t_c}) \;=\; \frac{1}{n}\sum_{i=1}^n |\,e^i - s_{es}^i\,|
\]
The optimal tuple is then chosen as
\[
\mathbf{t_c}^\star \;=\; \min_{\mathbf{t_{c}} \in \mathcal{E}_{\mathrm{EVL}}} \bar{D}(\mathbf{t_{c}})
\]
This criterion favors checkpoint tuples that are jointly loss-aligned and closest on average to the individual early-stopping epochs.

The purpose of this selection step is to identify the most representative checkpoint combination for model fusion. Each sensor’s early-stopping epoch $s^{es}_i$ corresponds to its best generalization point, as determined by validation performance. Selecting epochs too far from these points increases the risk that one or more sensors contribute checkpoints that are either overfitted (if much later than $s^{es}_i$) or underfitted (if much earlier). By minimizing the average epoch deviation $\bar{D}(\mathbf{t_c})$, we ensure that the selected tuple $\mathbf{t_c}^\star$ remains close to the generalization-optimal region for all sensors, thereby reducing the likelihood of combining mismatched or unstable model states. This provides a balanced and reliable basis for subsequent model merging.

\paragraph{Model fusion using the optimal tuple.}
Once the optimal loss-aligned tuple $\mathbf{t_c}^\star=(e_\star^1,e_\star^2,\dots,e_\star^n)$ is identified, we retrieve the corresponding checkpoints 
$\{\theta_i^{(e_\star^i)}\}_{i=1}^n$ from each sensor-specific model $\{\mathcal{M}_i\}_{i=1}^n$.  
These checkpoints represent mutually compatible convergence stages across sensors.

We then merge the selected models using a \emph{task arithmetic} algorithm, which operates directly on the model parameters to form a unified representation. The resulting merged model is denoted as

\[
\mathrm{\model} = \mathcal{T}\!\left(\, \theta_1^{(e_\star^1)}, \theta_2^{(e_\star^2)}, \dots, \theta_n^{(e_\star^n)} \,\right),
\]

\noindent where $\mathcal{T}$ indicates task arithmetic operation. Specifically, we employ the \textit{addition} operation defined in \cite{ilharco2022editing} to merge models.

This approach ensures that the fusion process integrates models trained to comparable validation performance levels and located near their respective generalization optima, thereby enhancing stability and reducing the risk of overfitted or underfitted contributions from individual sensors. We illustrate the intuition behind why EVL is more stable and generalizable compared to other checkpoint selection strategies in Section \ref{subsec:evl_intuition}.

\section{Pre-training Data}
\label{sec:pretraining_data}

\begin{figure*}[!htbp]
  \centering

    \makebox[0.31\textwidth][c]{\textbf{\footnotesize HiRISE}}%
    \makebox[0.31\textwidth][c]{\textbf{\footnotesize CTX}}%
    \makebox[0.31\textwidth][c]{\textbf{\footnotesize THEMIS}}\\[3pt]

  \begin{subfigure}{0.155\textwidth}
    \includegraphics[width=\linewidth]{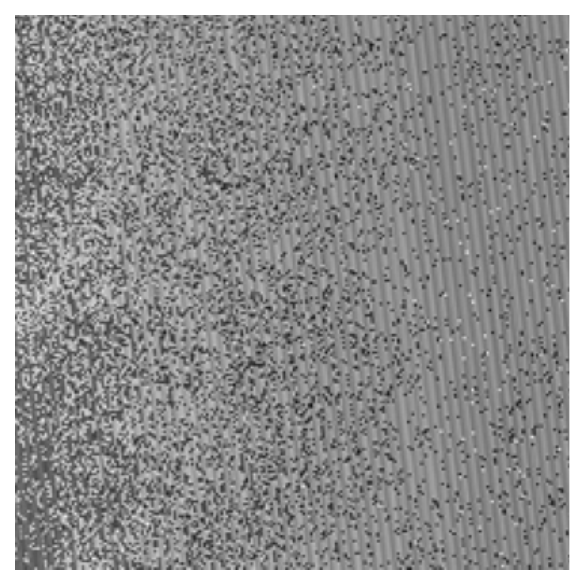}
  \end{subfigure}
  \begin{subfigure}{0.155\textwidth}
    \includegraphics[width=\linewidth]{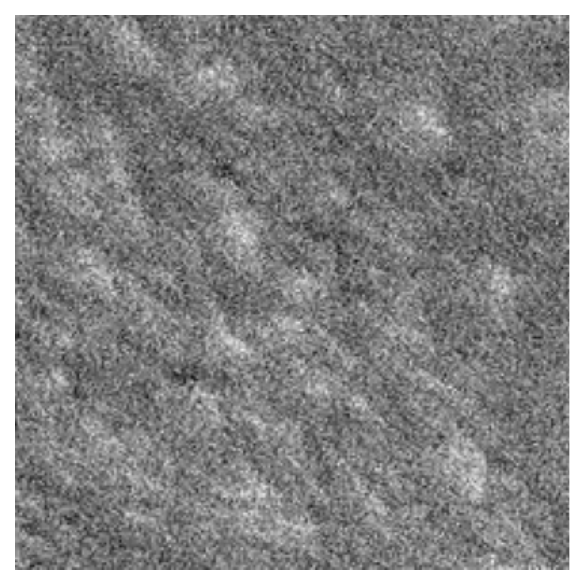}
  \end{subfigure}
  \begin{subfigure}{0.155\textwidth}
    \includegraphics[width=\linewidth]{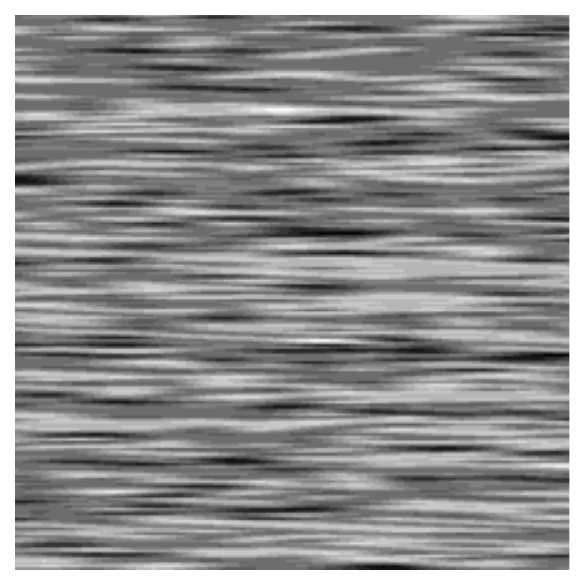}
  \end{subfigure}
  \begin{subfigure}{0.155\textwidth}
    \includegraphics[width=\linewidth]{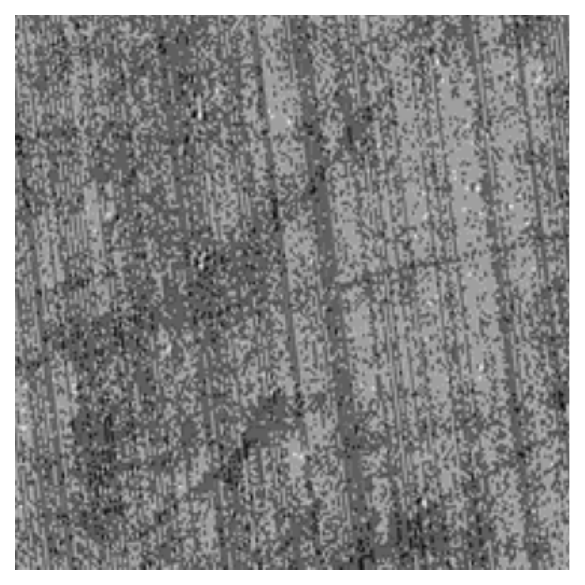}
  \end{subfigure}
  \begin{subfigure}{0.155\textwidth}
    \includegraphics[width=\linewidth]{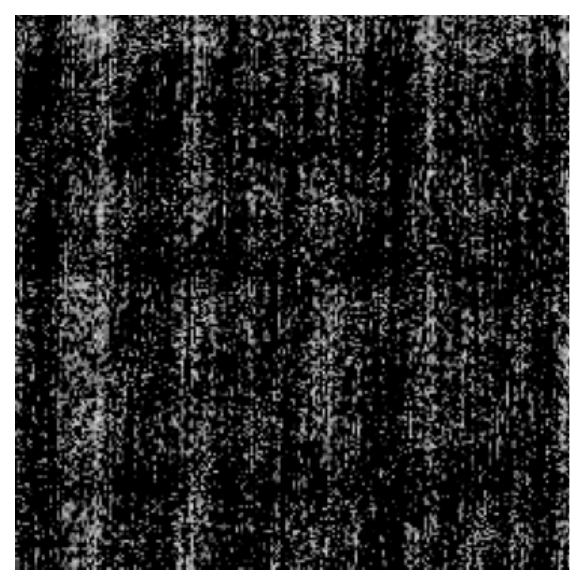}
  \end{subfigure}
  \begin{subfigure}{0.155\textwidth}
    \includegraphics[width=\linewidth]{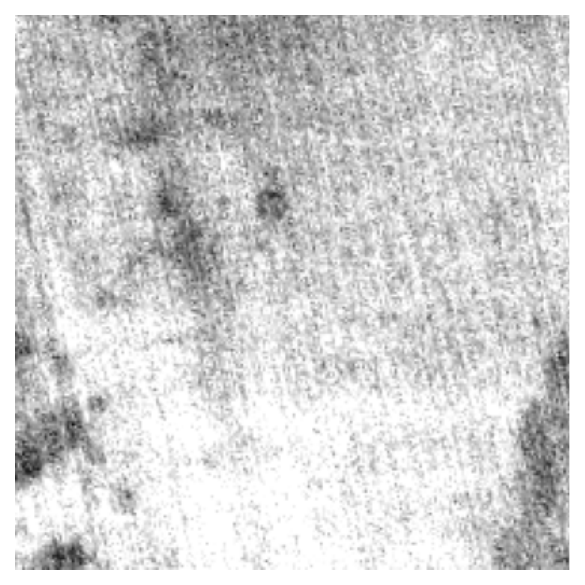}
  \end{subfigure}
  \\[4pt]

  \begin{subfigure}{0.155\textwidth}
    \includegraphics[width=\linewidth]{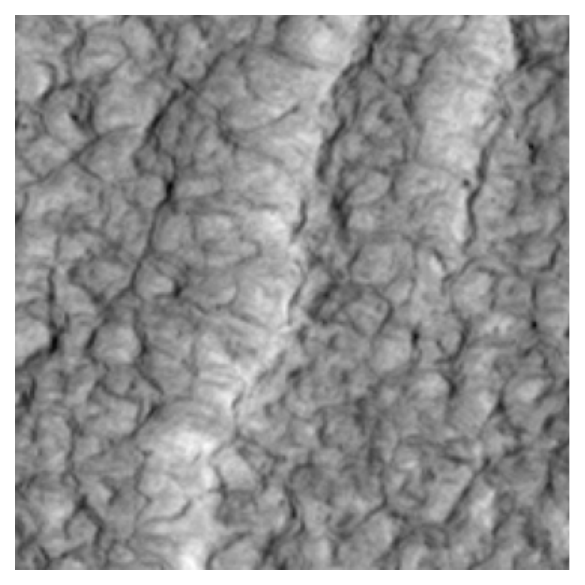}
  \end{subfigure}
  \begin{subfigure}{0.155\textwidth}
    \includegraphics[width=\linewidth]{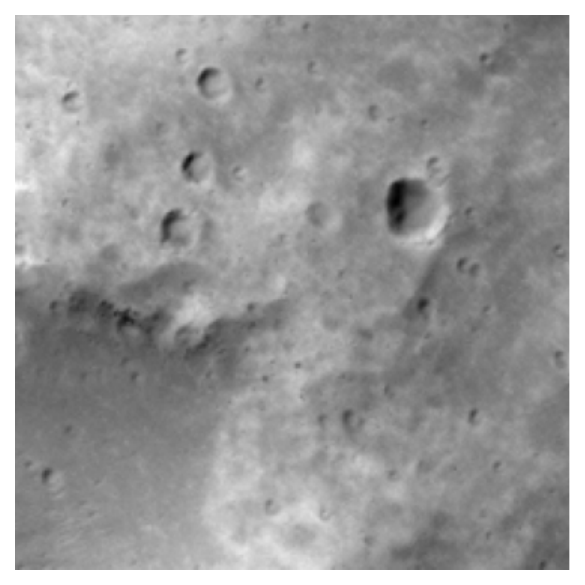}
  \end{subfigure}
  \begin{subfigure}{0.155\textwidth}
    \includegraphics[width=\linewidth]{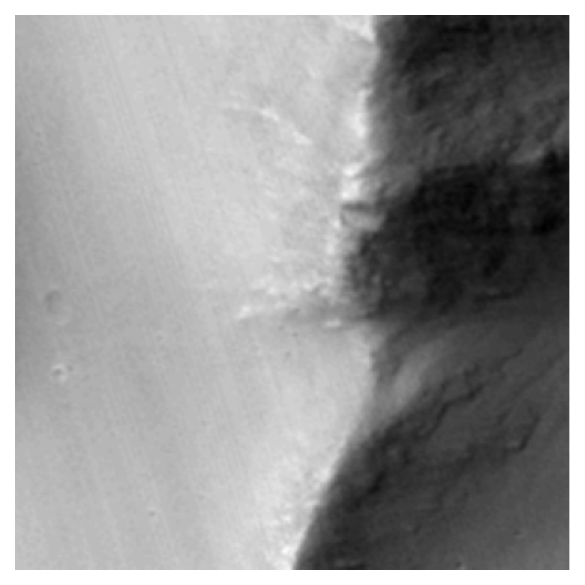}
  \end{subfigure}
  \begin{subfigure}{0.155\textwidth}
    \includegraphics[width=\linewidth]{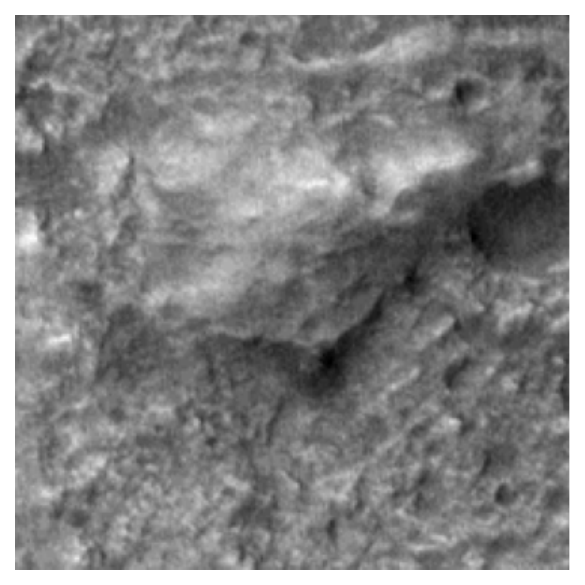}
  \end{subfigure}
  \begin{subfigure}{0.155\textwidth}
    \includegraphics[width=\linewidth]{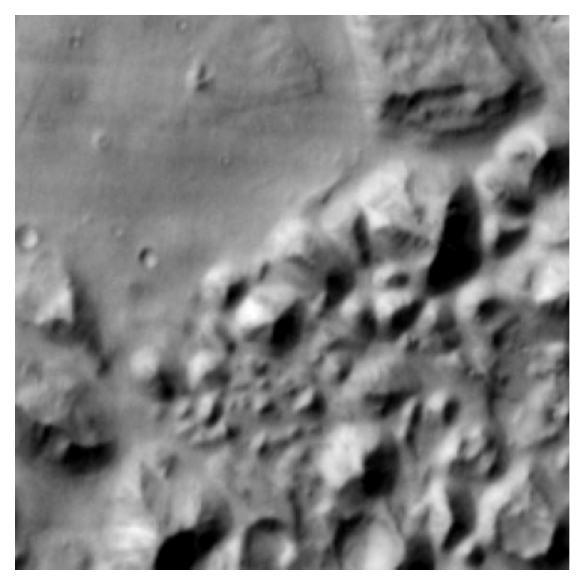}
  \end{subfigure}
  \begin{subfigure}{0.155\textwidth}
    \includegraphics[width=\linewidth]{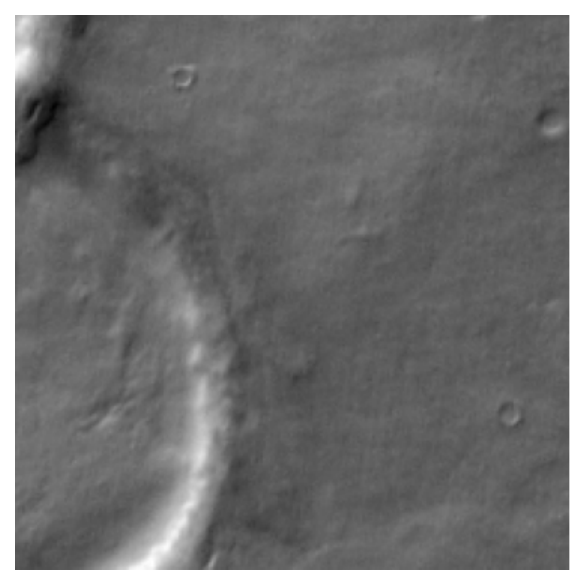}
  \end{subfigure}

  \caption{Illustrative samples of poor- and high-quality image samples from the HiRISE, CTX, and THEMIS sensors. The top row shows rejected low-quality samples exhibiting artifacts, blur, or noise, while the bottom row shows high-quality samples retained for pre-training.}
  \label{fig:data_samples}
\end{figure*}

As discussed in Section \ref{sec:introduction}, \model{} is an orbital foundation model trained on a large-scale, diverse dataset derived from multiple Martian sensors. Specifically, we utilize data from three key sources:

\begin{itemize}
    \item the High Resolution Imaging Science Experiment (HiRISE) \cite{mcewen2007mars} available at \textbf{0.25 m/pixels},
    \item the ConTeXt Camera (CTX) \cite{malin2007context, bell2013calibration} available at \textbf{5 m/pixels}, and
    \item the THermal EMission Imaging System (THEMIS) \cite{christensen2004thermal} available at \textbf{100 m/pixel}.
\end{itemize}

We select these sensors because all the orbital downstream tasks \cite{purohit2025marsbench} used and evaluated in this study belong to these three sensors. Details of the sensor types, characteristics, and the pre-training data preparation from all three sensors are provided in Appendix \ref{subsec:pretraining_data_details}.

As THEMIS is a low-resolution sensor, even with full surface coverage, we obtained a total of $\sim4M$ (millions) images from it. However, HiRISE and CTX contain $\sim16M$ and $\sim10M$ images, respectively. To ensure balanced representation, we sample 4M images from each sensor.

To include samples from a wide range of surface types and ensure that HiRISE and CTX retain their original distribution after downsampling to 4M, we proportionally sample data based on the geologic map units from the USGS Scientific Investigations Map 3292 — The Geologic Map of Mars (GMoM) \cite{tanaka2014geologic}. The GMoM divides Mars into 44 surface geologic units that represent a wide range of ages, morphologies, and compositions. We perform stratified sampling within each GMoM unit proportional to the unit’s area coverage for each of the three instrument datasets. This approach ensures that the final dataset not only balances sample counts across sensors but also preserves the geographic and geologic representativeness inherent to the original orbital coverage.

To ensure the quality and reliability of the pre-training corpus, we apply a filtering pipeline to remove low-quality samples containing satellite artifacts, noise, or blur. For each image from the three sensors, we compute two quantitative quality metrics:
\begin{itemize}
    \item Structural Similarity Index (SSIM) \cite{nilsson2020understanding}: We apply Gaussian smoothing to each image to reduce high-frequency noise while preserving structural content, and then compute the SSIM between the smoothed and original image.
    \item Noise Estimate \cite{liu2006noise}: The noise level is estimated by applying a Laplacian-like high-pass filter to emphasize intensity variations, followed by computing a statistical measure ($\sigma$) of the mean absolute filtered image.
\end{itemize}
Both metrics range from 0 to 1, where lower values indicate poor image quality. We discard samples with values below 0.4 (decided based on human verification) in both metrics for all sensors. This automated filtering step substantially reduces artifacts and ensures that only visually consistent, high-quality images contribute to pre-training. A few examples of good and poor-quality images are shown in Figure \ref{fig:data_samples}.

Finally, we split the data from all three sensors into training and validation sets using the HEALPix strategy \cite{gorski1999healpix}, which partitions a sphere into equal-area cells. These cells are then randomly divided into training and validation subsets. The same set of cells is used across all three sensors to prevent data leakage during model merging. The resulting global distribution and data splits for all three sensors are provided in Appendix \ref{subsec:pretraining_data_details}. We will publicly release this dataset as open-source to support future research in Mars science and foundation models.

\section{Experimental Framework}
\label{sec:experiments}

\subsection{Baselines}
\label{subsec:baselines}

We evaluate \model{} against a variety of baseline models to assess the effectiveness of our proposed approach. As discussed in Section \ref{sec:introduction}, ImageNet pre-trained models remain the default initialization strategy for many planetary and geospatial applications; therefore, we include an ImageNet pre-trained model as one of our baselines. We also consider a zero-shot setting with randomly initialized weights to quantify the contribution of pre-training, referred as \textit{scratch}.

Since no prior foundation models have been developed specifically for Mars, we further compare \model{} with several leading EO-FMs, including \textit{SatMAE}, \textit{CROMA}, \textit{Prithvi}, and \textit{TerraFM}. In addition, motivated by the strong representation quality demonstrated by \textit{DINOv3} in large-scale visual pre-training studies, we include its satellite pre-trained variant (trained on the SAT-493M dataset).

We also evaluate \textit{sensor-specific} pre-training baselines, where models are trained exclusively on data from a single sensor and then fine-tuned on corresponding downstream tasks. This allows us to analyze the relative benefits of same-sensor versus cross-sensor pre-training compared to \model{}. Furthermore, we consider a \textit{joint data pre-training} configuration, referred to as the \textbf{Data Merge (DM)} setting, in which data from all three sensors are directly combined to pre-train a single unified model.

Lastly, since we propose an optimal checkpoint selection strategy, we compare it against other checkpoint selection approaches. Specifically, we evaluate two alternatives: Early Stopping (ES) and Last Epoch (LE) merging. In the ES setting, models are merged using the early-stopping checkpoint from each sensor, while in the LE setting, models are merged using their final training checkpoint from each sensor.

\begin{table*}[]
\resizebox{\textwidth}{!}{
\begin{tabular}{l|cccc|ccccc|c}
\toprule[1.5pt]
& \textbf{AtmosDust}      & \textbf{DoMars16k} & \textbf{Frost}     & \textbf{Landmark}  & \textbf{Boulder} & \textbf{ConeQuest}  & \textbf{Crater Binary} & \textbf{Crater Multi} & \textbf{MMLS}      & {\begin{tabular}[c]{@{}l@{}}\textbf{Avg.}\\ \textbf{Rank} $\boldsymbol{\downarrow}$\end{tabular}} \\

\midrule[1pt]
Scratch & 
{0.94 $\pm$ 0.003} & {0.73 $\pm$ 0.008} & {0.95 $\pm$ 0.007} & {0.79 $\pm$ 0.010} & 
{0.07 $\pm$ 0.012} & {0.52 $\pm$ 0.035} & {0.37 $\pm$ 0.047} & 
{0.05 $\pm$ 0.017} & {\underline{0.50 $\pm$ 0.017}} & 
{4.11}
\\
ImageNet & 
{0.92 $\pm$ 0.010} & {0.91 $\pm$ 0.003} & {0.97 $\pm$ 0.009} & {\textbf{0.92 $\pm$ 0.003}} & 
{0.16 $\pm$ 0.034} & {\underline{0.70 $\pm$ 0.022}} & {\textbf{0.55 $\pm$ 0.012}} & 
{\underline{0.11 $\pm$ 0.007}} & {\textbf{0.57 $\pm$ 0.014}} & 
{\underline{2.33}}
\\
\midrule
DINOv3 & 
{\textbf{0.97 $\pm$ 0.000}} & {0.90 $\pm$ 0.000} & {\textbf{0.99 $\pm$ 0.013}} & {\underline{{0.91 $\pm$ 0.000}}} & 
{0.12 $\pm$ 0.106} & {0.51 $\pm$ 0.000} & {0.32 $\pm$ 0.000} & 
{0.01 $\pm$ 0.000} & {0.29 $\pm$ 0.000} & 
{3.67}
\\
CROMA & 
{0.83 $\pm$ 0.021} & {0.41 $\pm$ 0.010} & {0.76 $\pm$ 0.000} & {0.47 $\pm$ 0.000} & 
{0.17 $\pm$ 0.018} & {0.44 $\pm$ 0.000} & {0.27 $\pm$ 0.000} & 
{0.01 $\pm$ 0.001} & {0.14 $\pm$ 0.014} & 
{5.89}
\\
Prithvi & 
{0.81 $\pm$ 0.000} & {0.64 $\pm$ 0.008} & {0.81 $\pm$ 0.000} & {0.63 $\pm$ 0.000} & 
{0.04 $\pm$ 0.051} & {0.49 $\pm$ 0.000} & {0.18 $\pm$ 0.000} & 
{0.01 $\pm$ 0.000} & {0.23 $\pm$ 0.025} & 
{6.11}
\\
SatMAE & 
{\underline{0.96 $\pm$ 0.000}} & {\textbf{0.93 $\pm$ 0.000}} & {0.97 $\pm$ 0.000} & \textbf{{0.92 $\pm$ 0.000}} & 
{0.05 $\pm$ 0.000} & {0.68 $\pm$ 0.000} & {0.46 $\pm$ 0.000} & 
{0.04 $\pm$ 0.003} & {0.32 $\pm$ 0.000} & 
{3.00}
\\
TerraFM & 
{\textbf{0.97 $\pm$ 0.000}} & {0.89 $\pm$ 0.000} & {\textbf{0.99 $\pm$ 0.000}} & {0.86 $\pm$ 0.002} & 
{\textbf{0.21 $\pm$ 0.052}} & {0.38 $\pm$ 0.021} & {0.44 $\pm$ 0.000} & 
{0.04 $\pm$ 0.003} & {0.14 $\pm$ 0.103} & 
{3.67}
\\
\midrule
\rowcolor{lighttan}
MOMO & 
{\underline{0.96 $\pm$ 0.005}} & {\underline{0.92 $\pm$ 0.000}} & {\underline{0.98 $\pm$ 0.003}} & {\underline{0.91 $\pm$ 0.003}} & 
{\underline{0.20 $\pm$ 0.005}} & {\textbf{0.71 $\pm$ 0.008}} & {\underline{0.54 $\pm$ 0.005}} & 
{\textbf{0.12 $\pm$ 0.014}} & {\textbf{0.57 $\pm$ 0.009}} & 
{\textbf{1.67}}

\\
\bottomrule[1pt]
\end{tabular}
}
\caption{Performance comparison of different baselines with MOMO. Reported metrics include F1-Score for classification tasks and mIoU for segmentation tasks. \textbf{Bold} and \underline{Underlined} numbers indicate the best and second best performance.}
\label{tab:main_results}
\end{table*}

\subsection{Pre-training}
\label{subsec:pretraining}

We pre-train a separate model on data from each sensor for five epochs. As described in Section \ref{sec:pretraining_data}, each model is trained on $\sim4M$ samples. During training, we perform validation after every $\sim100k$ samples, recording the loss and saving a model checkpoint at each validation step. These intermediate checkpoints are later used for loss alignment and model merging procedures. Details about other hyperparameters are provided in Appendix \ref{sec:experimental_details}.

\subsection{Downstream Tasks}
\label{subsec:downstream_tasks}

We evaluate MOMO and baselines on all orbital tasks from Mars-Bench \cite{purohit2025marsbench}. For datasets where performance has already saturated, we present results in Appendix \ref{subsec:downstream_tasks_details} (two classification tasks). In the main paper, we report results for \textbf{four classification} tasks: \textit{AtmosDust}, \textit{DoMars16k}, \textit{Frost}, and \textit{Landmark}; and \textbf{five segmentation} tasks: \textit{Boulder}, \textit{ConeQuest}, \textit{Crater Binary}, \textit{Crater Multi}, and \textit{MMLS}. A concise summary and representative visual samples for all tasks are provided in Appendix \ref{subsec:downstream_tasks_details}. For each dataset–model combination, we perform hyperparameter tuning and report the best-performing configuration.

\section{Results and Analysis}
\label{sec:results_analysis}

\begin{table}[]
\resizebox{\columnwidth}{!}{
\begin{tabular}{l|ccc}
\toprule[1.5pt]
 \textbf{MOMO} & \textbf{Boulder} & \textbf{ConeQuest} & \textbf{Crater Binary} 
\\
\midrule[1pt]
ES & 
{0.12 $\pm$ 0.078} & {0.68 $\pm$ 0.0134} & {0.50 $\pm$ 0.014}
\\
LE &
{0.18 $\pm$ 0.031} & {0.70 $\pm$ 0.005} & {0.50 $\pm$ 0.015}
\\
\midrule
\rowcolor{lighttan}
EVL (Ours) & 
{\textbf{0.20 $\pm$ 0.005}} & {\textbf{0.71 $\pm$ 0.008}} & {\textbf{0.54 $\pm$ 0.005}}
\\
\bottomrule[1pt]
\end{tabular}
}
\caption{Performance comparison across different checkpoint selection strategies with the proposed EVL technique.}
\label{tab:checkpoint_results}
\end{table}

\begin{table*}[]
\resizebox{\textwidth}{!}{
\begin{tabular}{l|cccc|ccccc}
\toprule[1.5pt]
& \textbf{AtmosDust}      & \textbf{DoMars16k} & \textbf{Frost}     & \textbf{Landmark}  & \textbf{Boulder} & \textbf{ConeQuest}  & \textbf{Crater Binary} & \textbf{Crater Multi} & \textbf{MMLS} \\

\midrule[1pt]
HiRISE & 
\colorbox{blue!10}{{0.93 $\pm$ 0.018}} & {0.88 $\pm$ 0.002} & \colorbox{blue!10}{{0.97 $\pm$ 0.009}} & \colorbox{blue!10}{{0.90 $\pm$ 0.005}} & 
\colorbox{blue!10}{{0.12 $\pm$ 0.024}} & {0.66 $\pm$ 0.014} & {0.49 $\pm$ 0.017} & {0.06 $\pm$ 0.014} & {0.52 $\pm$ 0.040}
\\
CTX & 
{0.94 $\pm$ 0.012} & \colorbox{blue!10}{{0.90 $\pm$ 0.004}} & {0.95 $\pm$ 0.013} & {\textbf{0.91 $\pm$ 0.004}} & 
{0.17 $\pm$ 0.016} & \colorbox{blue!10}{{0.70 $\pm$ 0.015}} & {0.48 $\pm$ 0.008} & {0.07 $\pm$ 0.007} & \colorbox{blue!10}{{0.54 $\pm$ 0.008}}
\\

THEMIS & 
{0.94 $\pm$ 0.005} & {0.88 $\pm$ 0.003} & {0.94 $\pm$ 0.012} & {0.90 $\pm$ 0.004} & 
{0.17 $\pm$ 0.021} & {0.69 $\pm$ 0.024} & \colorbox{blue!10}{{0.50 $\pm$ 0.019}} & \colorbox{blue!10}{{0.07 $\pm$ 0.021}} & {0.51 $\pm$ 0.068}
\\
\midrule
DM & 
{0.94 $\pm$ 0.026} & {0.90 $\pm$ 0.003} & {0.44 $\pm$ 0.139} & {0.89 $\pm$ 0.004} & 
{0.14 $\pm$ 0.060} & {0.67 $\pm$ 0.012} & {0.52 $\pm$ 0.005} & {0.08 $\pm$ 0.019} & {0.48 $\pm$ 0.064}
\\
\midrule
\rowcolor{lighttan}
MOMO & 
{\textbf{0.96 $\pm$ 0.005}} & {\textbf{0.92 $\pm$ 0.002}} & {\textbf{0.98 $\pm$ 0.012}} & {\textbf{0.91 $\pm$ 0.003}} & 
{\textbf{0.20 $\pm$ 0.005}} & {\textbf{0.71 $\pm$ 0.008}} & {\textbf{0.54 $\pm$ 0.005}} & {\textbf{0.12 $\pm$ 0.012}} & {\textbf{0.57 $\pm$ 0.009}}

\\
\bottomrule[1pt]
\end{tabular}
}
\caption{Performance comparison across sensor-specific pre-training, joint data pre-training (DM - Data Merge), and \model. Here, \colorbox{blue!10}{highlighted} indicates the sensor associated with that downstream task. \textbf{Bold} number indicates the best performance.}
\label{tab:sensor_results}
\end{table*}

In this section, we present the results from all experiments, and compare MOMO against all baseline methods described in Section \ref{subsec:baselines}. The quantitative results are shown in Table \ref{tab:main_results}. For classification tasks, we report the weighted F1-score. And for segmentation tasks, we report mean Intersection over Union (mIoU). Since mIoU alone does not fully capture a model’s ability to accurately localize distinct features, we include the Object F1-Score as a complementary metric to provide a more comprehensive evaluation of segmentation performance.

\paragraph{Training from scratch and model pre-trained on natural images.} Table \ref{tab:main_results} demonstrates that directly training a model on downstream tasks performs worse than a pre-trained version of models. Comparing overall results with the ImageNet pre-trained version of the model, \model{} marginally outperforms ImageNet pre-training in the case of classification and segmentation tasks. For classification tasks, \model{} outperforms the ImageNet baseline, achieving an average improvement of $\sim1.25\%$ F1-score across all four tasks, achieving significant improvement ($\sim4\%$) for AtmosDust. ImageNet only outperforms in Landmark; however, the margin of improvement compared to \model{} is only 1\%. For segmentation, we also observe a similar trend where \model{} outperforms the ImageNet baseline, achieving an average improvement of $\sim1\%$ mIoU across all five tasks, achieving significant improvement ($\sim4\%$) for Boulder. These results indicate that while ImageNet pre-training provides comparable representations for classification, it fails to capture the spatial and textural characteristics required for precise feature localization.

\paragraph{Models Pre-trained on Earth Satellite data.} Consistent with the earlier observations, all EO-FMs perform on a similar level as MOMO in the case of classification tasks, but when comparing results in segmentation, overall MOMO outperforms compared to EO-FMs. DINOv3 shows competitive results for classification tasks.

\paragraph{Simple \textit{vs}. Complex Tasks.} Mars-Bench contains tasks of varying levels of complexity, which is also reflected in the results presented in Table \ref{tab:main_results}. For instance, in the classification category, binary classification datasets such as \textit{AtmosDust} and \textit{Frost} exhibit relatively higher performance, with F1-scores ranging from 0.96 to 0.98. In contrast, multi-class classification datasets such as \textit{DoMars16k} and \textit{landmark} yield slightly lower values, typically between 0.90 and 0.92. A similar trend is observed in segmentation tasks, where the \textit{Boulder} and \textit{MMLS} datasets differ significantly in performance. The \textit{Boulder} dataset achieves considerably lower scores (mIoU $\sim$ 0.04–0.21), whereas \textit{MMLS} obtains much higher values in some models (mIoU $\sim$ 0.14–0.57). This large performance gap primarily arises from the differences in object scale and count. In \textit{Boulder} segmentation, each image typically contains 25 or more small objects, making feature localization and boundary detection substantially more difficult. In contrast, \textit{MMLS} contains only one or two large landslides per image, with some occupying more than 50\% of the total area, allowing the model to learn their spatial features more easily and achieve higher mIoU. A similar pattern is also seen in the \textit{Crater Multi} and \textit{Crater Multi} datasets. While the model performs well in identifying crater regions (\textit{Crater Binary}), it struggles to classify them into their corresponding morphological types (\textit{Crater Multi}), resulting in a noticeable performance drop between the two tasks.

\paragraph{Comparison with Sensor-specific Pre-training.} We analyze the effect of fine-tuning performance under two settings: pre-training on the same sensor and pre-training on a different sensor (cross-sensor). The results are presented in Table \ref{tab:sensor_results}. It can be clearly observed that same-sensor pre-training consistently improves performance across all downstream tasks compared to cross-sensor pre-training. However, cross-sensor pre-training does not lead to a significant drop in performance, indicating reasonable generalization across sensors. When compared with \textbf{MOMO}, it achieves an improvement of 1.75\% in classification and 4.2\% in segmentation. Although the gain in classification is relatively small, MOMO offers a key advantage over sensor-specific training. Specifically, sensor-specific approaches require maintaining separate models for each sensor, whereas MOMO provides a unified model that can handle all sensors simultaneously.

\paragraph{Comparison with Joint Data Pre-training.} We compare MOMO with Data Merge (DM) approach. Table~\ref{tab:sensor_results} shows that MOMO outperforms baselines by 15.25\% in classification and 5.4\% in segmentation, demonstrating its consistently strong performance across tasks. Also, the DM approach suffers from significant limitations in scalability and modularity. Specifically, if additional data or tasks from a new sensor become available in the future, DM would require complete re-training using data aggregated from all sensors. In contrast, \model{} supports easy integration by pre-training a model only on the new sensor and then merging it with existing sensor-specific models using the same EVL-based merging framework. This design makes \model{} more flexible, computationally efficient, and generalizable across new sensors and modalities.

\begin{figure*}[htbp]
    \centering
    \begin{subfigure}[b]{0.3\textwidth}
        \includegraphics[width=\linewidth]{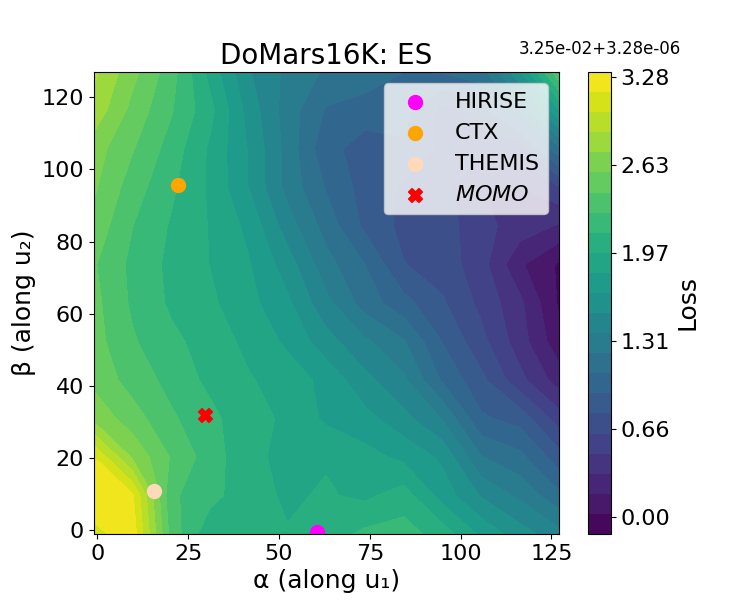}
    \end{subfigure}
    \begin{subfigure}[b]{0.3\textwidth}
        \includegraphics[width=\linewidth]{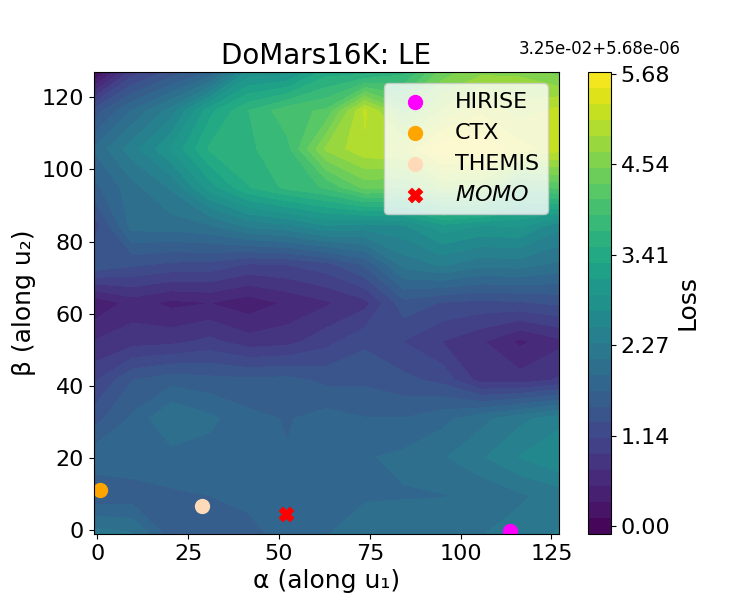}
    \end{subfigure}
    \begin{subfigure}[b]{0.3\textwidth}
        \includegraphics[width=\linewidth]{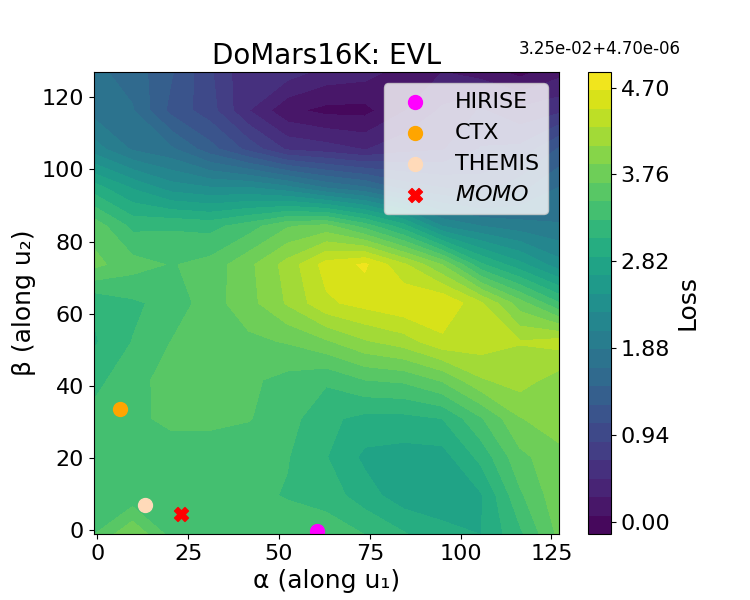}
    \end{subfigure}

    \begin{subfigure}[b]{0.3\textwidth}
        \includegraphics[width=\linewidth]{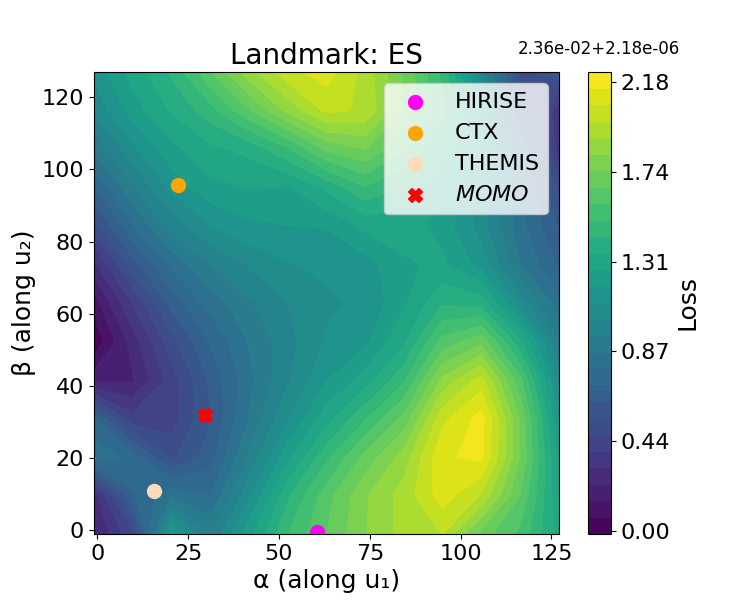}
    \end{subfigure}
    \begin{subfigure}[b]{0.3\textwidth}
        \includegraphics[width=\linewidth]{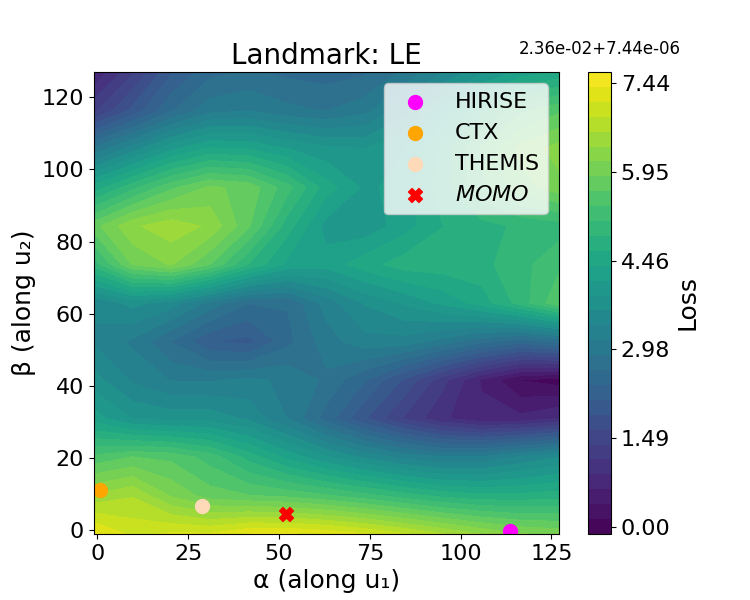}
    \end{subfigure}
    \begin{subfigure}[b]{0.3\textwidth}
        \includegraphics[width=\linewidth]{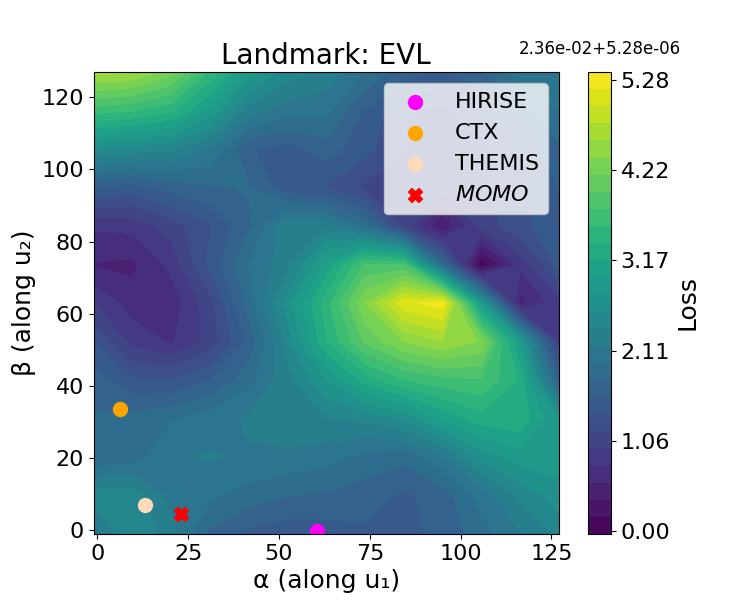}
    \end{subfigure}

    \caption{Loss landscape visualization across different checkpoint selection strategies on DoMars16k and Landmark datasets. The red markers represent MOMO obtained using Early Stopping (ES), Last Epoch (LE), and Equal Validation Loss (EVL), respectively.}
    \label{fig:loss_landscape}
\end{figure*}

\paragraph{Comparison of Checkpoint Selection Strategies.} To demonstrate the effectiveness of our proposed EVL-based checkpoint selection strategy, we evaluate downstream performance using different checkpoint selection methods. Since classification tasks generally show minimal variation across models, this analysis is conducted only on segmentation tasks, selecting one downstream task from each sensor. As shown in Table \ref{tab:checkpoint_results}, the EVL-based MOMO consistently achieves the best overall performance with average $\sim2.5\%$ mIoU improvement across three tasks. The only exception is observed in the \textit{ConeQuest} dataset, where the difference remains negligible ($1\%$) compared to the Early Stopping (ES) and Last Epoch (LE) strategies. Moreover, the definition of the “last epoch” itself is ambiguous, as it may occur at any iteration depending on the training setup (e.g., epoch 5 or 500), making it an unreliable criterion for checkpoint selection. These results highlight that EVL provides a more stable approach for merging models trained across different sensors.

\subsection{Intuition Behind EVL}
\label{subsec:evl_intuition}

To further show the efficiency of the optimal checkpoint strategy, we conduct experiments where we examine the loss landscape resulting from various model-merging strategies. To visualize this landscape, we adopt the projection method used by \citet{garipov2018loss, wortsman2022model}, which maps models onto a two-dimensional plane. This method employs Gram-Schmidt orthogonalization to identify two orthogonal directions in the weight space, defining the x- and y-axes of the landscape illustrated in Figure \ref{fig:loss_landscape}. The basis vectors for the orthogonalization are derived from the ImageNet, HiRISE, and CTX pre-trained models. The parameters $\alpha$ and $\beta$ denote unit displacements along the x and y axes, respectively.

As it can be observed from Table \ref{tab:main_results}, binary classification tasks do not show significant performance differences (only 1-2\% different), we select \textit{DoMars16k} and \textit{Landmark} datasets for this analysis. The losses are computed on the \textit{DoMars16k} and \textit{Landmark} datasets using class-balanced versions of their respective test sets.

In Figure \ref{fig:loss_landscape}, each plot presents the interpolated loss surface among the HiRISE, CTX, and THEMIS models. Across all merging strategies, the task-vector-based merged model consistently achieves an equal or lower loss compared to its constituent models. This demonstrates that task-vector merging can produce models that outperform the originals from which they were derived.

Prior work \cite{ainsworth2022git, gargiulo2025task, qu2024rethinking} suggests that optimal merging occurs when constituent models lie within the same loss basin, as this promotes stability and enhanced performance. Figure \ref{fig:loss_landscape} shows that relative to both LE and ES, the EVL strategy selects model checkpoints that are more closely aligned in weight space. Consequently, EVL is expected to yield the most stable and best-performing merged model.

\section{Conclusions}
\label{sec:conclusion}

In this work, we introduced \model, the first foundation model designed for Mars orbital applications. \model{} effectively integrates multi-sensor and multi-resolution data through a model-merging framework that leverages our proposed optimal checkpoint selection strategy. This approach ensures stable and compatible model fusion, allowing the model to generalize efficiently across diverse data sources. Trained on a large-scale corpus of $\sim12$ million curated samples from HiRISE, CTX, and THEMIS sensors, \model{} demonstrates strong performance across all Mars-Bench orbital downstream tasks. Findings from experimental results indicate that \model{} outperforms compared to ImageNet pre-training, EO-FMs, sensor-specific training, data merge approach, and different checkpoint selection strategies. Particularly, \model{} demonstrates consistently superior performance in segmentation tasks, which require precise identification of fine-grained details within images. Moreover, our approach offers a scalable training framework that enables efficient integration of new sensors without requiring complete re-training. In summary, \model{} represents the first foundation model in planetary science, and we believe this effort will inspire the development of future foundation models for planetary science research.

\vspace{-4.2mm}

\paragraph{Limitations} Our approach builds on a model-merging framework; however, due to computational constraints, we did not include additional model-merging-based baselines for comparison. We also do not explore alignment-based techniques in this work, which can be used in conjunction with our method and may further improve performance, as suggested in literature \cite{zhang2025beyond, theus2025generalized, ainsworth2023git}. Finally, our method relies on the assumption of linear mode connectivity between models. While this assumption holds in many practical settings, it may not strictly apply when models are trained on highly divergent data distributions or when models are functionally equivalent but differ due to network symmetries.

\newpage

\noindent \textbf{Acknowledgment:} Part of this research was carried out at the Jet Propulsion Laboratory, California Institute of Technology, under a contract with the National Aeronautics and Space Administration.

{
    \small
    \bibliographystyle{ieeenat_fullname}
    \bibliography{main}
}


\clearpage
\onecolumn

\appendix

\vspace*{10mm}

\section{Data Overview}
\label{sec:data}

\subsection{Pre-training Data Details}
\label{subsec:pretraining_data_details}

\begin{wrapfigure}{r}{0.25\linewidth}
    \centering
    \includegraphics[width=0.14\textwidth]{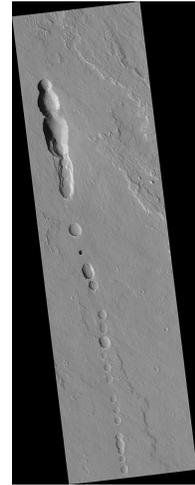}
    \caption{Example of a HiRISE map-projected image used in our study. The dark border around the image represents no-data regions that were filtered out during preprocessing to ensure high-quality crop selection.}
    \label{fig:hirise_samples}
\end{wrapfigure}

\paragraph{HiRISE} is mounted on the Mars Reconnaissance Orbiter (MRO) satellite and has been collecting data since 2006. HiRISE captures visible spectrum images at very high-resolution, i.e., $\sim0.25$ meters/pixel. HiRISE images cover a cumulative area of $\sim4.5\%$ of the martian surface; however, unique coverage (excluding repeats for stereo and monitoring) is $<3\%$ \cite{mcewen2024high}. We used grayscale data from the RED band of map-projected Reduced Data Record (RDR) products, and from the Primary and Extended Science Phases (PSP and ESP)\footnote{\url{https://hirise-pds.lpl.arizona.edu/PDS/RDR/}}. Our square image crops were extracted from map-projected HiRISE images. We applied a filter to exclude crops that extended into the no-data HiRISE border (black area in the Figure \ref{fig:hirise_samples}). We gathered $\sim16M$ image crops, which were selected from images acquired between November 2006 through May 2025. From these, we first filter the data using SSIM and Noise Estimate, and then further downsample to $\sim4M$ using GMOM stratified sampling as described in Section \ref{sec:pretraining_data}. We adopt GMOM-based sampling instead of random sampling to ensure uniform geographic coverage, as random sampling may miss certain regions of the surface. As shown in prior work \cite{purohit2025how, Plekhanova_2025_CVPR}, geographic distribution plays an important role in model performance.

\paragraph{CTX} is another visible imager on MRO with a wider ground footprint. To prepare pre-training data for CTX, we used open-source CTX data from the Murray Lab\footnote{\url{https://murray-lab.caltech.edu/CTX/tiles/beta01/}} (updated March 2023) \cite{murray_lab}. The dataset is a seam-corrected global image mosaic of Mars rendered at $5.0$ meters/pixel \cite{malin2007context, dickson2023release}. Data covers the entirety of the Martian surface ($> 99.5\%$). The global image data is divided into 3960 geotiff tiles (4$^{\circ}$ $\times$ 4$^{\circ}$) from 88$^{\circ}$S to 88$^{\circ}$N \cite{dickson2018global, dickson2023release}. Each tile is subdivided into four subtiles (2$^{\circ}$ $\times$ 2$^{\circ}$). On the Murray Lab, CTX data was last updated in March 2023. To create almost even geographic distribution from all subtiles, in each subtile, we randomly sample 630 points and crop data samples. This way, we make sure that we are capturing the diversity of the terrain across the Martian surface. This resulted in $\sim10M$ CTX data samples globally, and then we filter this data to remove noisy samples (using SSIM and Noise Estimate). From there, we further sample $\sim4M$ data samples using GMOM as described in Section \ref{sec:pretraining_data}.

\paragraph{THEMIS} is a thermal infrared imager on the Mars Odyssey Orbiter and has been collecting data since 2001. We used THEMIS day-time images at 100 meters/pixel resolution. THEMIS has global coverage \cite{christensen2004thermal}. Similar to HiRISE data, original THEMIS tiles are tilted. Thus, we have used the same process (as HiRISE) to create crops from THEMIS tiles as well. We have used Projected Brightness Temperatures (PBT) products from THEMIS archive\footnote{\url{https://static.mars.asu.edu/pds/ODTGEO_v2/data/}} \cite{christensen2001mars}. Although THEMIS has global coverage, due to low-resolution data, we got a total of $\sim4$M data samples. We have exported and processed data from October 2002 to April 2025.

\begin{figure}[htbp]
    \centering
    \includegraphics[width=0.9\textwidth]{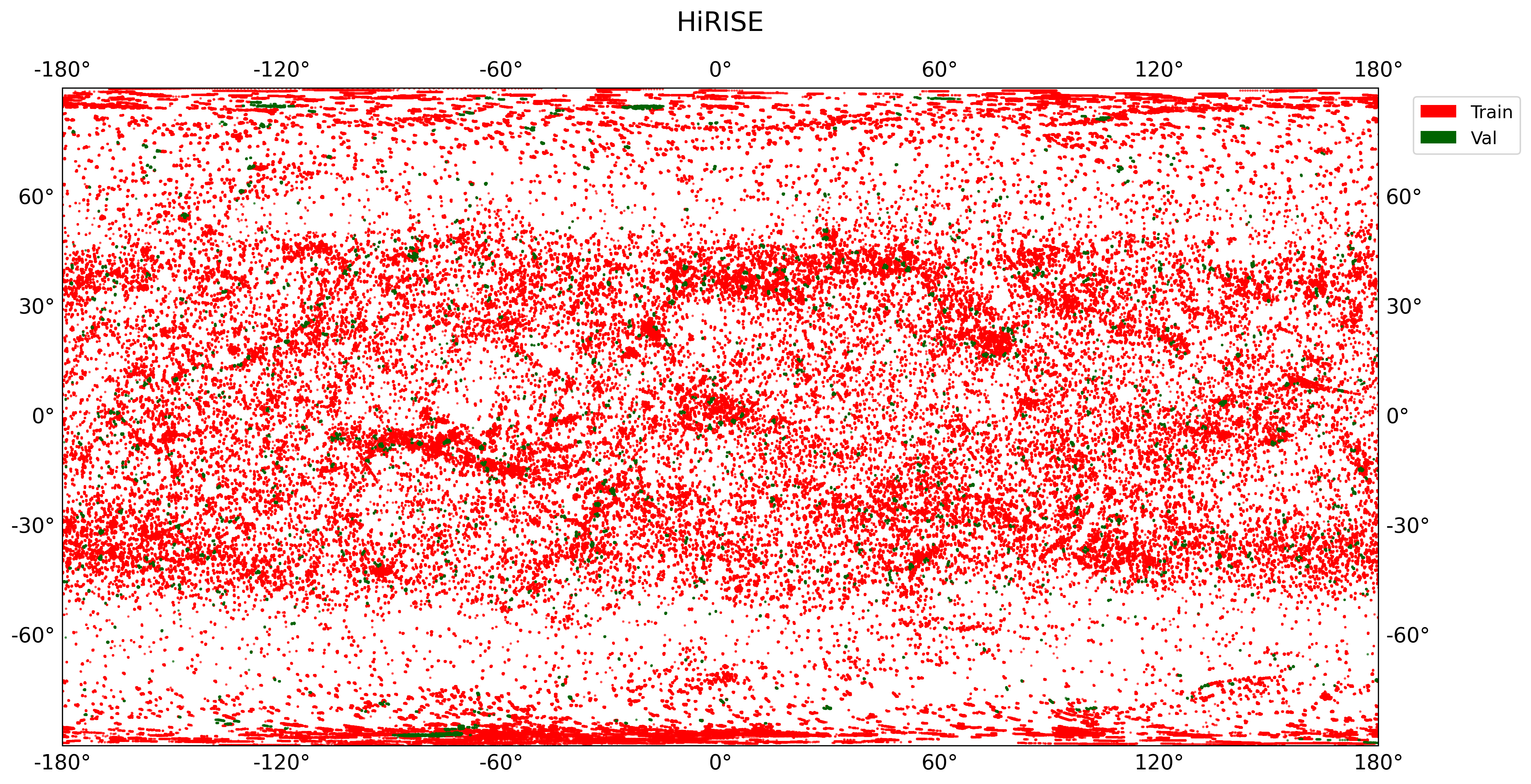}
    \caption{HiRISE pre-training data distribution}
    \label{fig:hirise_pretraining}
\end{figure}

\begin{figure}[htbp]
    \centering
    \includegraphics[width=0.9\textwidth]{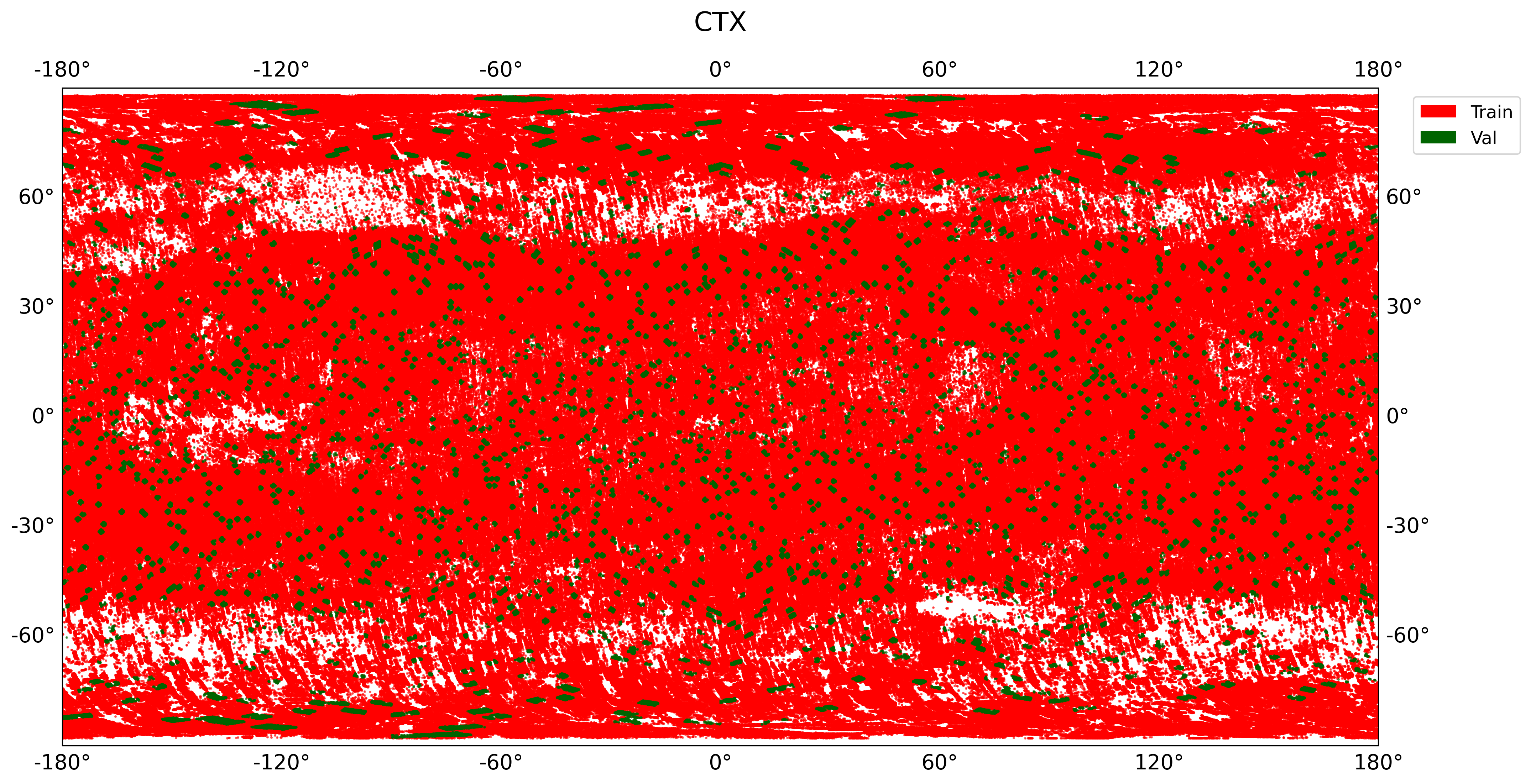}
    \caption{CTX pre-training data distribution}
    \label{fig:ctx_pretraining}
\end{figure}

\begin{figure}[htbp]
    \centering
    \includegraphics[width=0.9\textwidth]{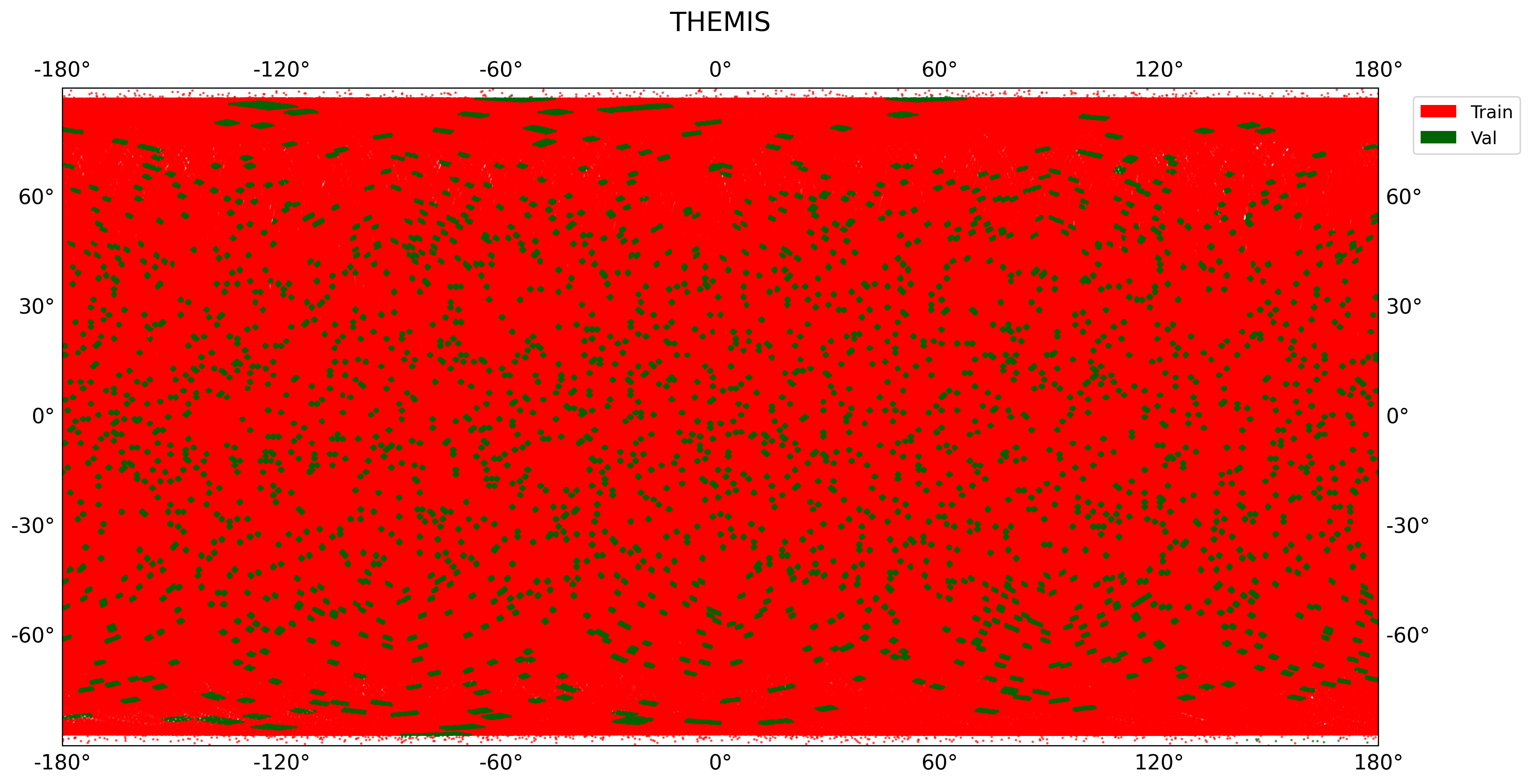}
    \caption{THEMIS pre-training data distribution}
    \label{fig:themis_pretraining}
\end{figure}

As described in Section \ref{sec:pretraining_data}, we use a HEALPix strategy to create geographically consistent training and validation sets. We use a HEALPix pixel size of 64, ensuring that all samples within a given cell are assigned exclusively to either the training or validation split. From our $\sim4M$ curated samples, we split 95\% for training and 5\% for validation for each sensor, respectively. This prevents cross-sensor leakage and preserves geographic diversity within each split. The resulting spatial distribution and the final train/validation assignments for HiRISE, CTX, and THEMIS are summarized in Figure~\ref{fig:hirise_pretraining}, Figure~\ref{fig:ctx_pretraining}, and  Figure~\ref{fig:themis_pretraining}, respectively.

\newpage
\null
\newpage

\subsection{Downstream Tasks}
\label{subsec:downstream_tasks_details}

As mentioned in Section \ref{sec:experiments}, we evaluate MOMO on all orbital tasks from Mars-Bench \cite{purohit2025marsbench}. In this section, we describe details of each downstream task and which sensor that downstream task belongs to. For simplicity, we remove the prefix ``mb-" from all datasets, and for long dataset names, we represent that with a short, meaningful name.

\subsubsection{Classification}

\begin{figure}[htbp]

  \centering
  \begin{subfigure}{\columnwidth}
    \centering
    \includegraphics[width=0.2\columnwidth]{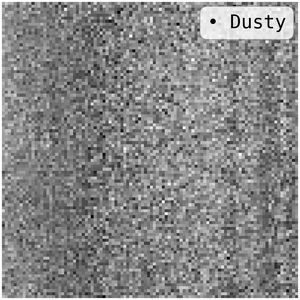}
    \includegraphics[width=0.2\columnwidth]{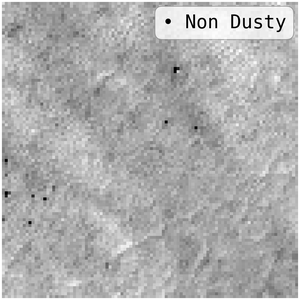}
  \end{subfigure}

  \caption{AtmosDust}
  \label{fig:atmospheric_dust_cls}

\end{figure}

\paragraph{AtmosDust} This is a binary classification dataset and focuses on classifying between ``\textbf{Dusty}'' and ``\textbf{Non dusty}'' regions in Mars surface imagery captured by the HiRISE sensor on the MRO. This dataset has two versions provided in Mars-Bench, i.e., EDR (Experimental Data Record) and RDR (Reduced Data Record). As both datasets have the same characteristics, we have evaluated only on the RDR version of the dataset (Figure \ref{fig:atmospheric_dust_cls}). The \texttt{EDR} refers to raw images from the sensor that have not been calibrated or stitched together; while the \texttt{RDR} is a downsampled or processed version of the EDR, typically used for quick viewing or initial analysis.

\begin{figure*}[htbp]

  \centering
  \begin{subfigure}{\columnwidth}
    \centering
    \includegraphics[width=0.13\columnwidth]{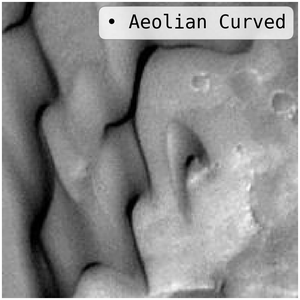}
    \includegraphics[width=0.13\columnwidth]{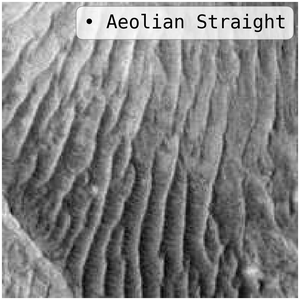}
    \includegraphics[width=0.13\columnwidth]{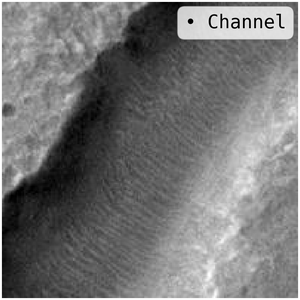}
    \includegraphics[width=0.13\columnwidth]{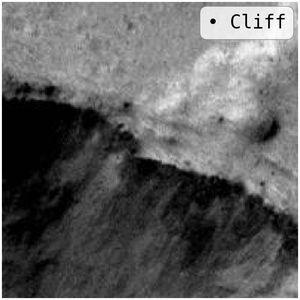}
    \includegraphics[width=0.13\columnwidth]{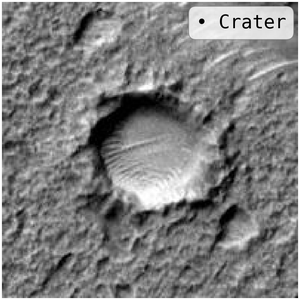}
  \end{subfigure}

  \centering
  \begin{subfigure}{\columnwidth}
    \centering
    \centering
    \includegraphics[width=0.13\columnwidth]{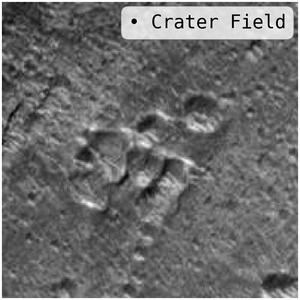}
    \includegraphics[width=0.13\columnwidth]{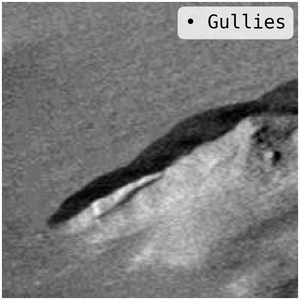}
    \includegraphics[width=0.13\columnwidth]{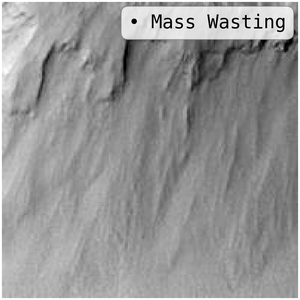}
    \includegraphics[width=0.13\columnwidth]{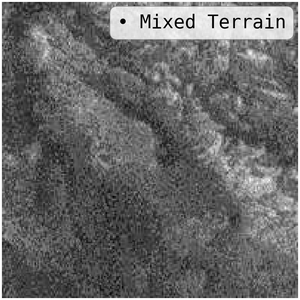}
    \includegraphics[width=0.13\columnwidth]{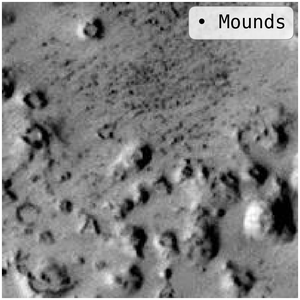}
  \end{subfigure}

  \centering
  \begin{subfigure}{\columnwidth}
    \centering
    \centering
    \includegraphics[width=0.13\columnwidth]{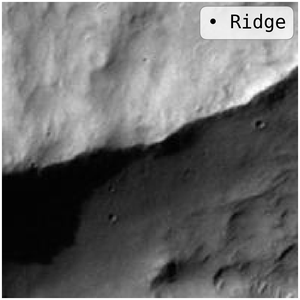}
    \includegraphics[width=0.13\columnwidth]{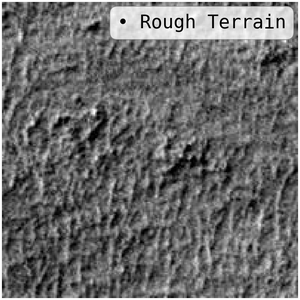}
    \includegraphics[width=0.13\columnwidth]{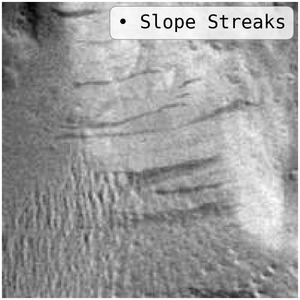}
    \includegraphics[width=0.13\columnwidth]{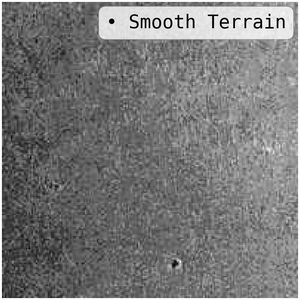}
    \includegraphics[width=0.13\columnwidth]{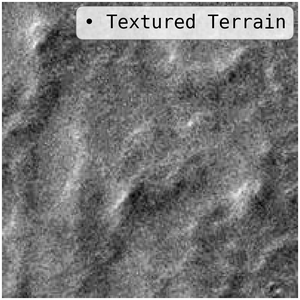}
  \end{subfigure}

  \caption{DoMars16k}
  \label{fig:domars16k}

\end{figure*}

\paragraph{DoMars16k} This is a multi-class classification dataset designed for geomorphologic feature recognition on Mars using imagery from the \textit{CTX} sensor. It consists of 15 classes (Figure \ref{fig:domars16k}) grouped into five thematic categories: (1) \textbf{Aeolian Bedforms:} Aeolian Curved, Aeolian Straight; (2) \textbf{Topographic Landforms:} Channel, Cliff, Mounds, Ridge; (3) \textbf{Slope Features:} Gullies, Mass Wasting, Slope Streaks; (4) \textbf{Impact Landforms:} Crater, Crater Field; and (5) \textbf{Basic Terrain:} Mixed Terrain, Rough Terrain, Smooth Terrain, Textured Terrain. This is one of the largest and most diverse \textit{orbital} datasets in terms of $\#$ of classes. Hence, the dataset presents a unique challenge due to its class granularity, significant variability within classes, and subtle differences between classes, making it valuable for evaluating models.

\begin{figure*}[htbp]

  \centering
  \begin{subfigure}{\columnwidth}
    \centering
    \includegraphics[width=0.13\columnwidth]{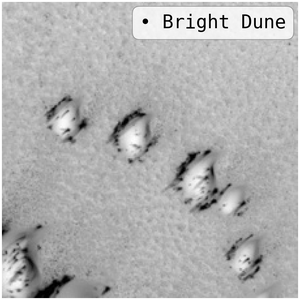}
    \includegraphics[width=0.13\columnwidth]{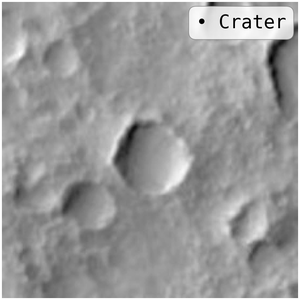}
    \includegraphics[width=0.13\columnwidth]{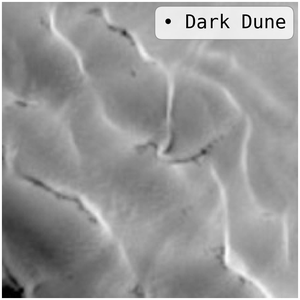}
    \includegraphics[width=0.13\columnwidth]{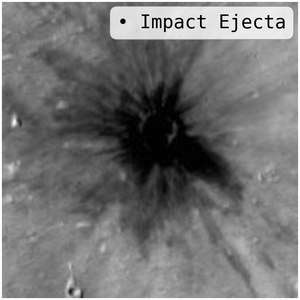}
  \end{subfigure}

  \centering
  \begin{subfigure}{\columnwidth}
    \centering
    \centering
    \includegraphics[width=0.13\columnwidth]{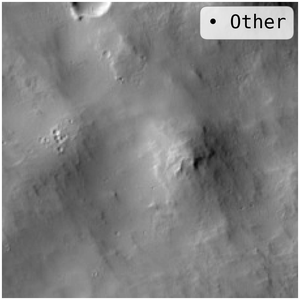}
    \includegraphics[width=0.13\columnwidth]{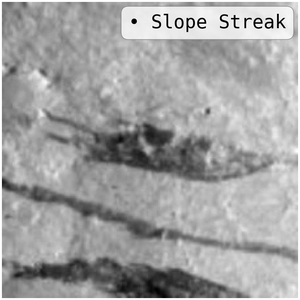}
    \includegraphics[width=0.13\columnwidth]{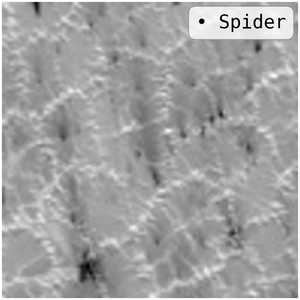}
    \includegraphics[width=0.13\columnwidth]{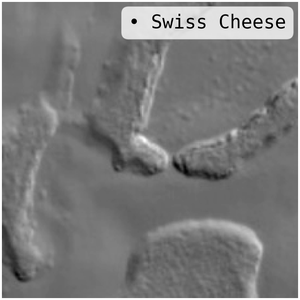}
  \end{subfigure}

  \caption{Landmark}
  \label{fig:landmark_cls}

\end{figure*}

\paragraph{Landmark} This dataset is a multi-class classification corpus derived from orbital HiRISE imagery. Each image is assigned to one of eight geomorphological feature classes: \textbf{Bright Dune, Crater, Dark Dune, Impact Ejecta, Slope Streak, Spider, Swiss Cheese}, and \textbf{Other} (Figure~\ref{fig:landmark_cls}). The class distribution is highly imbalanced, with \textit{Other} dominating the dataset and \textit{Impact Ejecta} representing the rarest (minority) class.

\begin{figure}[htbp]

  \centering
  \begin{subfigure}{\columnwidth}
    \centering
    \includegraphics[width=0.2\columnwidth]{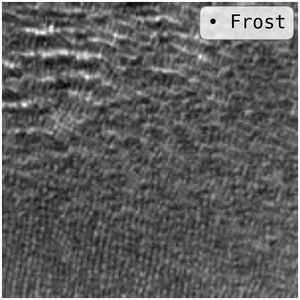}
    \includegraphics[width=0.2\columnwidth]{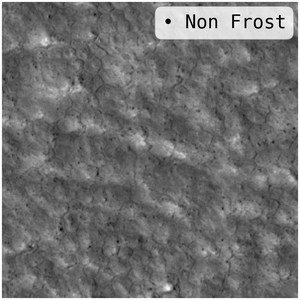}
  \end{subfigure}

  \caption{Frost}
  \label{fig:frost_cls}

\end{figure}

\paragraph{Frost} This is a binary classification dataset designed to detect the presence or absence of surface frost in Mars satellite imagery. The dataset consists of \textit{HiRISE} images labeled as either ``\textbf{Frost}'' or ``\textbf{Non Frost}'' (Figure \ref{fig:frost_cls}). Among all datasets in Mars-Bench, this is the largest in terms of the $\#$ of samples, and the dataset is well-balanced in terms of class distribution.

\paragraph{Saturated Task} Apart from the tasks described above, we exclude the mb-change\_cls task from our study, as both of its available versions, HiRISE and CTX, are already saturated. In prior benchmarks and in MOMO, this task consistently reaches near-perfect performance. Although the task exists in both HiRISE and CTX variants, the CTX version additionally suffers from an insufficient number of test samples for statistically meaningful evaluation. For completeness, we only evaluate the mb-change\_cls\_hirise dataset, but we do not include it in our core experiments.

\begin{figure}[htbp]
  \centering

  \begin{subfigure}[b]{0.48\textwidth}
    \centering
    \includegraphics[width=0.40\linewidth]{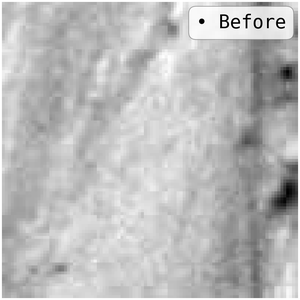}
    \includegraphics[width=0.40\linewidth]{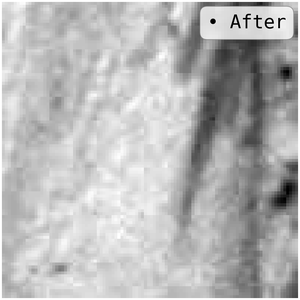}
    \caption{Change}
    \label{fig:change_hirise}
  \end{subfigure}
  \hspace{-10mm}
  \begin{subfigure}[b]{0.48\textwidth}
    \centering
    \includegraphics[width=0.40\linewidth]{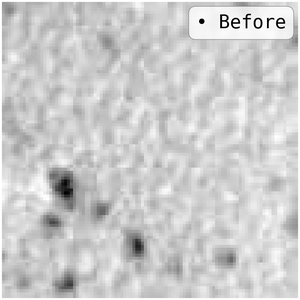}
    \includegraphics[width=0.40\linewidth]{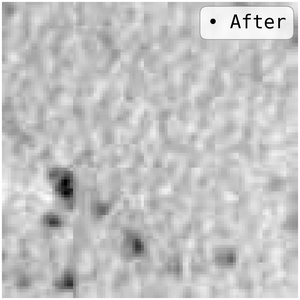}
    \caption{No change}
    \label{fig:no_change_hirise}
  \end{subfigure}

  \caption{mb-change\_cls\_hirise}
  \label{fig:change_cls_hirise}
\end{figure}

\textit{mb-change\_cls\_hirise} This dataset is designed for binary classification of surface changes using temporal image pairs; specifically, one image taken before and another after some time period, from the \textit{same} Martian location. The task involves identifying whether meaningful surface change has occurred and classifying between ``\textbf{Change}'' and ``\textbf{No change}''. Unlike standard single-image classification, this task requires forming a composite input from two grayscale images (Figure \ref{fig:change_cls_hirise}). Following the approach outlined by \citet{kerner2019toward}, we adopt the composite grayscale method: the blue channel encodes the ``before'' image, the green channel encodes the ``after'' image, and the red channel is set to zero.

For the \textit{mb-change\_cls\_hirise} dataset, we conducted experiments using MOMO and all baseline models, excluding EO-FMs and DINOv3. All models achieved 100\% accuracy and F1-score, indicating that the task is already saturated. Therefore, we did not include these results in the main paper and did not perform further experiments on EO-FMs for this dataset.

\subsubsection{Segmentation}

\begin{figure*}[htbp]
  \centering

  \begin{subfigure}[b]{0.48\textwidth}
    \centering
    \includegraphics[width=0.40\linewidth]{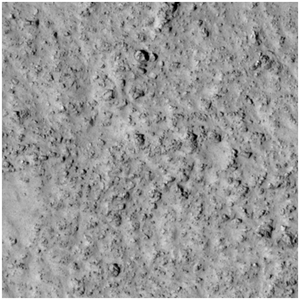}
    \includegraphics[width=0.40\linewidth]{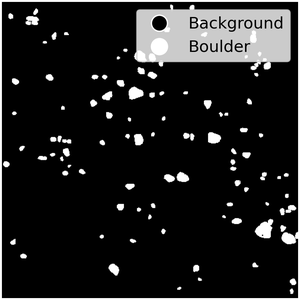}
  \end{subfigure}
  \hspace{-10mm}
  \begin{subfigure}[b]{0.48\textwidth}
    \centering
    \includegraphics[width=0.40\linewidth]{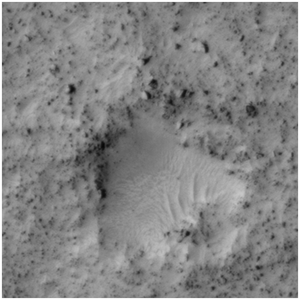}
    \includegraphics[width=0.40\linewidth]{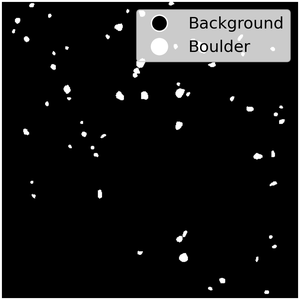}
  \end{subfigure}

  \caption{Boulder}
  \label{fig:boulder_seg}
\end{figure*}

\paragraph{Boulder} This is a binary segmentation dataset focused on segmenting boulders on the Martian surface using high-resolution orbital imagery from the HiRISE sensor. The dataset comprises manually annotated binary masks indicating the presence or absence of boulders within each image (Figure \ref{fig:boulder_seg}). Boulders were annotated by planetary scientists using precise polygon outlines, ensuring high-quality labels. This is one of the smallest datasets in Mars-Bench, with only tens of samples (i.e., 39), and that makes it challenging for the computer vision community.

\begin{figure*}[htbp]
  \centering

  \begin{subfigure}[b]{0.48\textwidth}
    \centering
    \includegraphics[width=0.40\linewidth]{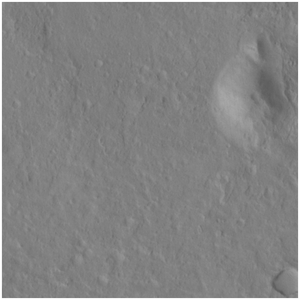}
    \includegraphics[width=0.40\linewidth]{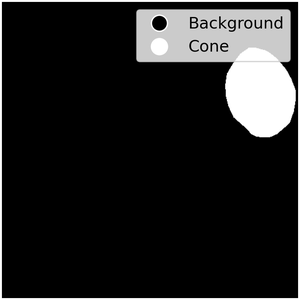}
  \end{subfigure}
  \hspace{-10mm}
  \begin{subfigure}[b]{0.48\textwidth}
    \centering
    \includegraphics[width=0.40\linewidth]{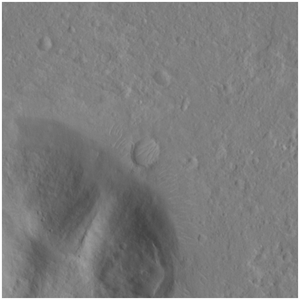}
    \includegraphics[width=0.40\linewidth]{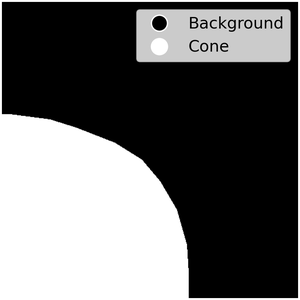}
  \end{subfigure}

  \caption{ConeQuest}
  \label{fig:conequest_seg}
\end{figure*}

\paragraph{ConeQuest} This is a binary segmentation dataset focused on identifying volcanic cones on the Martian surface using CTX imagery. It was developed to support global mapping and morphologic analysis of small-scale volcanic landforms. The dataset spans three geographically diverse regions on Mars, capturing substantial variation in cone shape, size, and appearance, making it a challenging benchmark for model generalization. Each sample consists of an image and its corresponding binary mask (Figure \ref{fig:conequest_seg}), with all annotations created and validated by expert geologists to ensure scientific accuracy. Particularly, the dataset includes negative samples (images without any cones), which introduces additional complexity by requiring models to correctly predict true negatives rather than detecting cones in every image.

\begin{figure*}[htbp]
  \centering

  \begin{subfigure}[b]{0.48\textwidth}
    \centering
    \includegraphics[width=0.40\linewidth]{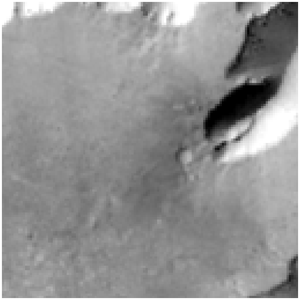}
    \includegraphics[width=0.40\linewidth]{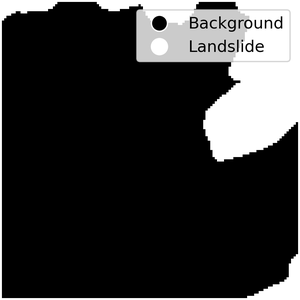}
  \end{subfigure}
  \hspace{-10mm}
  \begin{subfigure}[b]{0.48\textwidth}
    \centering
    \includegraphics[width=0.40\linewidth]{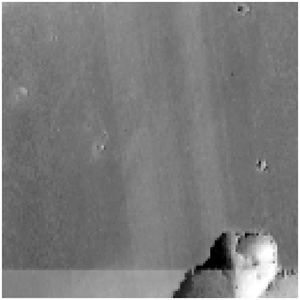}
    \includegraphics[width=0.40\linewidth]{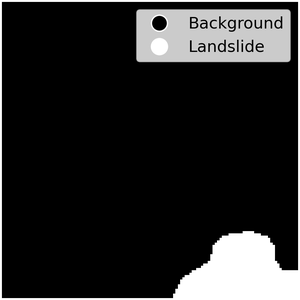}
  \end{subfigure}

  \caption{MMLS}
  \label{fig:mb-mmls}
\end{figure*}

\paragraph{MMLS} This is a binary segmentation dataset designed to identify landslides on the Martian surface, with a focus on the Valles Marineris region from the CTX sensor. All annotations were manually created by expert geologists, ensuring high-quality, scientifically accurate labels. Each image sample includes multi-modal satellite data comprising 7 channels: RGB (3), Digital Elevation Model (DEM), thermal inertia, slope, and grayscale intensity (Figure \ref{fig:mb-mmls} visualizes grayscale channels only). This rich set of modalities captures the complex geomorphology of landslide-prone regions, making the dataset especially valuable for developing and benchmarking robust segmentation models in planetary science. All experiments in this paper utilize only the RGB channels for training and evaluation.

\begin{figure*}[htbp]
  \centering

  \begin{subfigure}[b]{\textwidth}
    \centering
    \begin{subfigure}[b]{0.48\textwidth}
      \centering
      \includegraphics[width=0.40\linewidth]{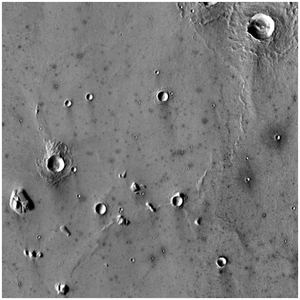}
      \includegraphics[width=0.40\linewidth]{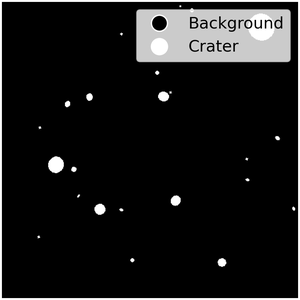}
    \end{subfigure}
    \hspace{-10mm}
    \begin{subfigure}[b]{0.48\textwidth}
      \centering
      \includegraphics[width=0.40\linewidth]{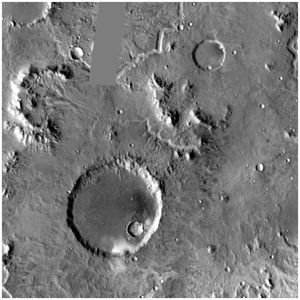}
      \includegraphics[width=0.40\linewidth]{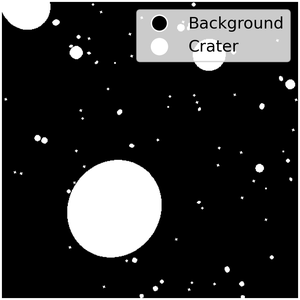}
    \end{subfigure}
    \caption{Crater Binary}
    \label{fig:crater_binary_seg}
  \end{subfigure}

  \vspace{5mm}

  \begin{subfigure}[b]{\textwidth}
    \centering
    \begin{subfigure}[b]{0.48\textwidth}
      \centering
      \includegraphics[width=0.40\linewidth]{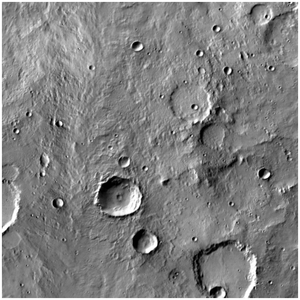}
      \includegraphics[width=0.40\linewidth]{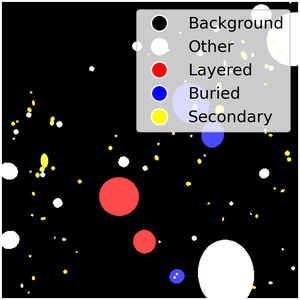}
    \end{subfigure}
    \hspace{-10mm}
    \begin{subfigure}[b]{0.48\textwidth}
      \centering
      \includegraphics[width=0.40\linewidth]{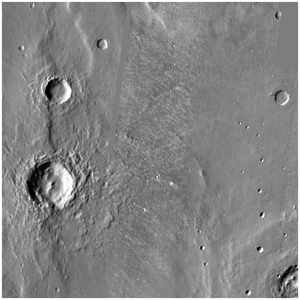}
      \includegraphics[width=0.40\linewidth]{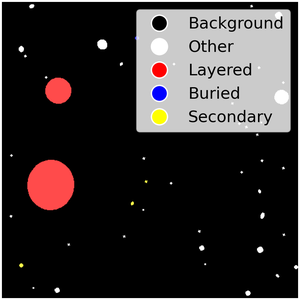}
    \end{subfigure}
    \caption{Crater Multi}
    \label{fig:crater_multi_seg}
  \end{subfigure}

  \caption{Crater Segmentation Datasets}
  \label{fig:crater_seg}
\end{figure*}

\paragraph{Crater Binary \& Crater Multi} These two datasets focus on crater segmentation using THEMIS imagery. In particular, mb-crater\_binary\_seg is a binary segmentation dataset that distinguishes \textbf{crater} vs. \textbf{non-crater} regions, while mb-crater\_multi\_seg is a multi-class segmentation dataset with four crater types: \textbf{Other}, \textbf{Layered}, \textbf{Buried}, and \textbf{Secondary} (Figure \ref{fig:crater_seg}).

\section{Experiments Details}
\label{sec:experimental_details}

\paragraph{Pre-training Experiments.} All pre-training experiments are conducted on the ViT-Base model on a single NVIDIA A100 GPU with a batch size of 256 at the JPL computing infrastructure. We apply only a random horizontal flip as data augmentation during training and use no augmentation for validation. Models are pre-trained with a learning rate of $10^{-3}$, a weight decay of $0.05$, a patch size of $16$, and a mask ratio of $0.75$. For each sensor-specific dataset, we train the model for $5$ epochs. We record the model state and loss values after every $100k$ processed samples, enabling consistent comparison of validation loss across all individually pre-trained models. During pre-training, all loss weights $\lambda_i$ are set to $0.25$, ensuring equal weightage to pixel-based and perceptual loss. For loss alignment, we use a patience of $5$ and a tolerance parameter of $\epsilon = 10^{-4}$. We analyze the effect of different values of the tolerance parameter in Section \ref{subsec:epsilon}. During model merging, we apply a scaling coefficient of $0.3$, following the recommendation of \citet{ilharco2022editing}. We further analyze the sensitivity of our approach to different scaling coefficients in Section \ref{subsec:scaling_coefficient}. For the Data Merge experiments, we apply the same hyperparameter configuration. For the ImageNet-pretrained baseline, we use the model provided by He et al. \cite{he2022masked}. During pre-training, a ViT-Base model requires approximately 12 hours to train on $\sim$4M samples for each individual sensor. In contrast, pre-training a ViT-Base model using the Data Merge ($\sim 12M$ data samples in pre-training) setup takes approximately 35 hours. 

\paragraph{Downstream Tasks Experiments.} For all downstream classification and segmentation tasks, we perform extensive hyperparameter tuning for each model–dataset combination. For classification, a linear layer is applied on top of the pre-trained encoder, whereas segmentation uses a U-NetFormer decoder. All classification datasets use cross-entropy loss, while segmentation employs a weighted combination of Dice, cross-entropy, and boundary losses. Because certain datasets are highly imbalanced (e.g., Landmark), we apply dataset-specific balancing strategies: no balancing for AtmosDust and Frost (nearly balanced), loss reweighting for DoMars16k, and oversampling for Landmark. For all segmentation tasks, we adopt loss reweighting, as background pixels dominate the ground-truth masks.

All models are trained for up to 100 epochs with an early-stopping patience of ${5, 10}$. We perform a sweep over hyperparameters: learning rates $\in {1\times10^{-3},1\times10^{-4}}$, weight decays $\in {5\times10^{-2}, 1\times10^{-1}}$, layer decays $\in {0.5, 0.6, 0.75}$, and warm-up epochs $\in {0, 5, 10}$. For segmentation, the loss-weighting coefficients are tuned using two settings: $(\text{Dice}, \text{CE}, \text{Boundary}) = (0.5, 0.2 ,0.3)$ and $(0.3, 0.5, 0.2)$.

For the DINOv3 model, we use the variant pre-trained on Earth satellite data, specifically the SAT-493M dataset. For the remaining EO-FMs, most do not provide an end-to-end fine-tuning reference codebase for downstream tasks, so we implement our own framework for both classification and segmentation.

To ensure robustness, we run each experiment five times with different random seeds and report the mean and standard deviation. All downstream experiments are conducted on A100 GPUs on ASU \cite{jennewein2023sol} or JPL servers, depending on GPU availability.

\section{Extended Results}
\label{sec:extended_results}

In this section, we present additional experiments and analyses that complement the results discussed in the main paper. These include the effect of model size, detailed evaluations of reconstruction quality, the influence of the scaling coefficient, comparison with the model currently deployed in the NASA PDS system, and examples demonstrating MOMO’s capability for generating global maps.

\subsection{Effect of Model Size}
\label{subsec:vit_variant}

\begin{table*}[htbp]
\resizebox{\textwidth}{!}{
\begin{tabular}{l|cccc|ccccc}
\toprule[1.5pt]
\textbf{MOMO} & \textbf{AtmosDust}      & \textbf{DoMars16k} & \textbf{Frost}     & \textbf{Landmark}  & \textbf{Boulder} & \textbf{ConeQuest}  & \textbf{Crater Binary} & \textbf{Crater Multi} & \textbf{MMLS}  
\\


\midrule[1pt]
ViT-Small
 & \textbf{0.96}         & 0.92    & 0.96  & 0.92          
 & \textbf{0.22} & 0.71 & 0.54 & 0.09 & 0.58         
\\
ViT-Base
& \textbf{0.96} & \textbf{0.93} & \textbf{0.97}  & \textbf{0.93} & 
0.18 & 0.72 & 0.56 & 0.12  & 0.58 
\\
ViT-Large
 & \textbf{0.96} & 0.92 & 0.96 & \textbf{0.93} & 
0.19 & \textbf{0.73} & \textbf{0.58} & \textbf{0.14} & \textbf{0.60} 
\\       
\bottomrule[1pt]
\end{tabular}
}
\caption{Performance comparison of ViT variants. Reported metrics include F1-Score for classification tasks, and mIoU for segmentation tasks. \textbf{Bold} numbers indicate the highest value in each column.}
\label{tab:vit_variants}
\end{table*}

To evaluate the robustness of our proposed approach across different model capacities, we conducted experiments using three Vision Transformer (ViT) variants: ViT-Small, ViT-Base, and ViT-Large. Each variant was pre-trained and evaluated under the same setup across all downstream tasks to examine how model size influences performance. The results are summarized in Table \ref{tab:vit_variants}.

From the results, we observe that in classification tasks, the performance difference across all three ViT variants is negligible, typically less than 1\%. However, in segmentation tasks, increasing model size clearly improves performance, with ViT-Large achieving the best results in most cases. An exception is observed in the \textit{Boulder} dataset, where ViT-Small outperforms larger models. This can be attributed to the small size of the dataset and the limited number of samples per class, which may lead to overfitting in larger models. Overall, these results indicate that while classification remains largely invariant to model capacity, segmentation benefits significantly from increased model size.

\subsection{Reconstruction}
\label{subsec:reconstruction}

\begin{figure*}[htbp]
\centering

\begin{subfigure}{0.14\textwidth}
    \centering
    \caption*{\textbf{Original}}
  \end{subfigure}
  \begin{subfigure}{0.14\textwidth}
    \centering
    \caption*{\textbf{HiRISE}}
  \end{subfigure}
  \begin{subfigure}{0.14\textwidth}
    \centering
    \caption*{\textbf{CTX}}
  \end{subfigure}
  \begin{subfigure}{0.14\textwidth}
    \centering
    \caption*{\textbf{THEMIS}}
  \end{subfigure}
  \begin{subfigure}{0.13\textwidth}
    \centering
    \caption*{\textbf{HCT}}
  \end{subfigure}
  
\begin{subfigure}{0.9\textwidth}
\centering
\includegraphics[width=0.15\textwidth]{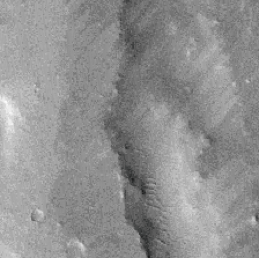}\hspace{1mm}%
\includegraphics[width=0.15\textwidth]{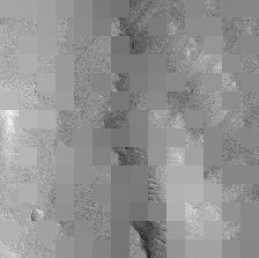}\hspace{1mm}%
\includegraphics[width=0.15\textwidth]{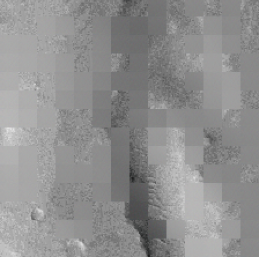}\hspace{1mm}%
\includegraphics[width=0.15\textwidth]{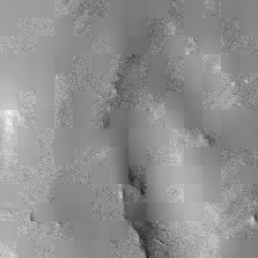}\hspace{1mm}%
\includegraphics[width=0.15\textwidth]{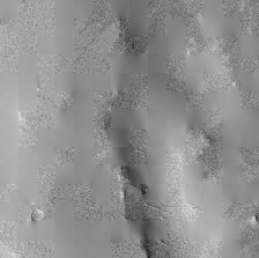}\\
[3pt]
\includegraphics[width=0.15\textwidth]{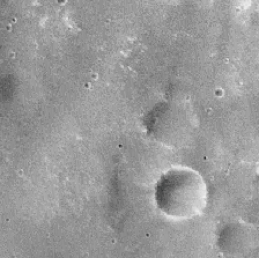}\hspace{1mm}%
\includegraphics[width=0.15\textwidth]{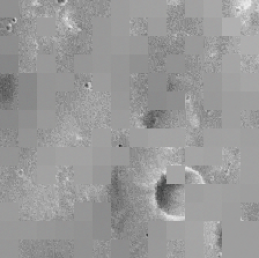}\hspace{1mm}%
\includegraphics[width=0.15\textwidth]{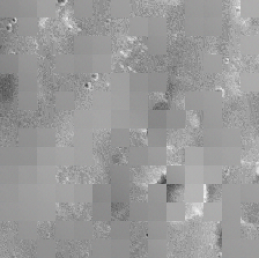}\hspace{1mm}%
\includegraphics[width=0.15\textwidth]{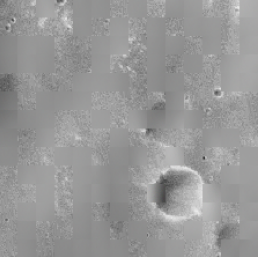}\hspace{1mm}%
\includegraphics[width=0.15\textwidth]{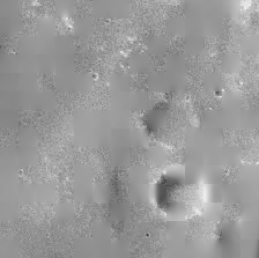}
\end{subfigure}

\caption{Reconstruction results using ViT-Base models pre-trained with only MSE loss. The figure compares the Original image against reconstructions from sensor-specific models (HiRISE, CTX, THEMIS) and the Data Merge model (HCT). The top row displays a HiRISE sample, and the second row displays a CTX sample.}
\label{fig:mse_reconstruction}
\end{figure*}

As described in Section \ref{sec:experimental_details}, our pre-training objective combines pixel-based loss with a perceptual loss. In this section, we evaluate the impact of this formulation by comparing it against a baseline that uses only MSE loss. Figure \ref{fig:mse_reconstruction} illustrates reconstruction results when ViT-Base is pre-trained on each sensor independently as well as using the Data Merge approach. We show one randomly selected HiRISE sample (top row) and one CTX sample (bottom row). Under the MSE-only objective, several patches are poorly reconstructed: the model often recovers the overall surface tone but fails to regenerate fine-scale geomorphological features. For example, in the CTX example (second row), when $\sim20\%$ of the crater is masked, the model reconstructs the surrounding terrain reasonably well but is unable to recover the crater structure itself.

\newpage

\begin{figure*}
\centering

\begin{subfigure}{0.13\textwidth}
    \centering
    \caption*{\textbf{Original}}
  \end{subfigure}
  \begin{subfigure}{0.13\textwidth}
    \centering
    \caption*{\textbf{Masked}}
  \end{subfigure}
  \begin{subfigure}{0.13\textwidth}
    \centering
    \caption*{\textbf{HiRISE}}
  \end{subfigure}
  \begin{subfigure}{0.13\textwidth}
    \centering
    \caption*{\textbf{CTX}}
  \end{subfigure}
  \begin{subfigure}{0.13\textwidth}
    \centering
    \caption*{\textbf{THEMIS}}
  \end{subfigure}
  \begin{subfigure}{0.13\textwidth}
    \centering
    \caption*{\textbf{HCT}}
  \end{subfigure}
  
\begin{subfigure}{0.9\textwidth}
\centering
\includegraphics[width=0.15\textwidth]{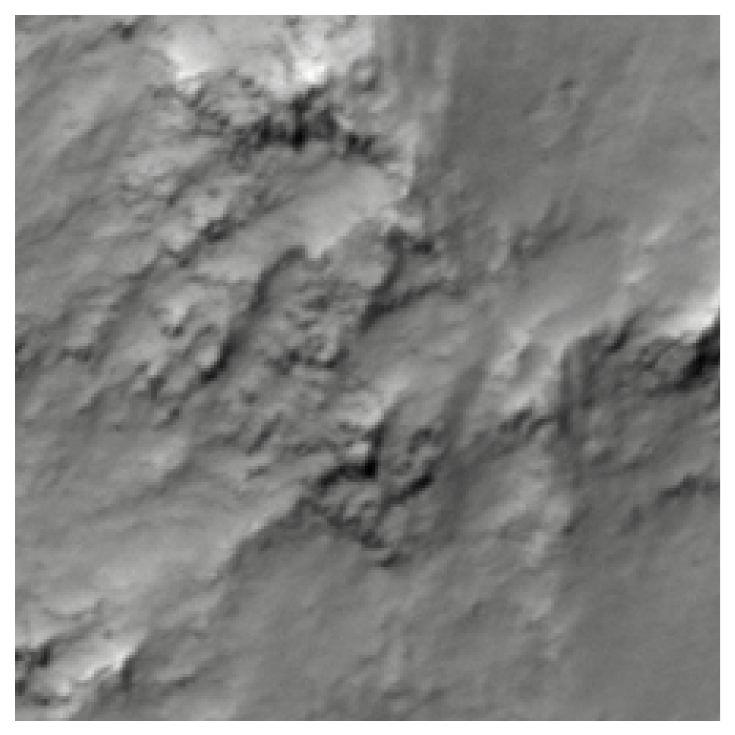}%
\includegraphics[width=0.15\textwidth]{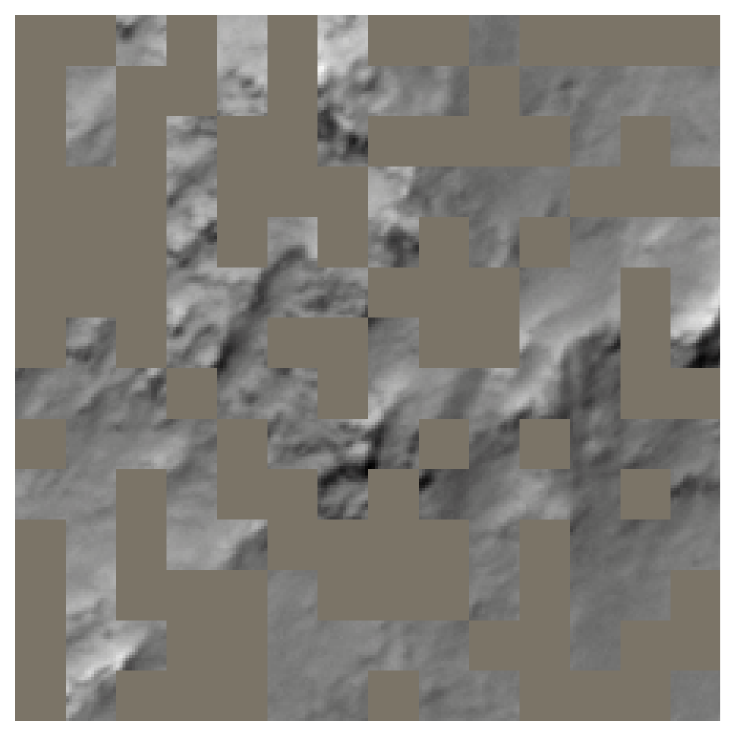}%
\includegraphics[width=0.15\textwidth]{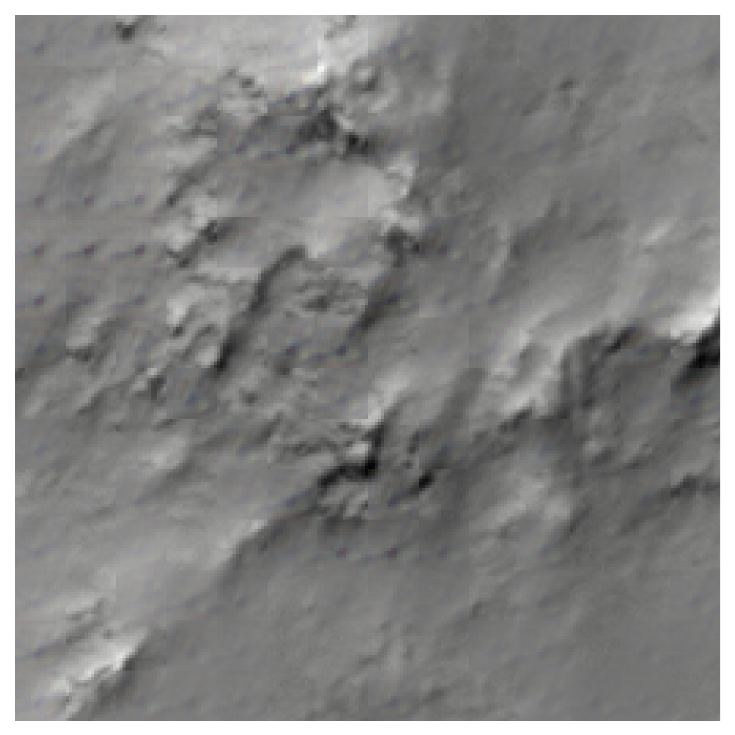}%
\includegraphics[width=0.15\textwidth]{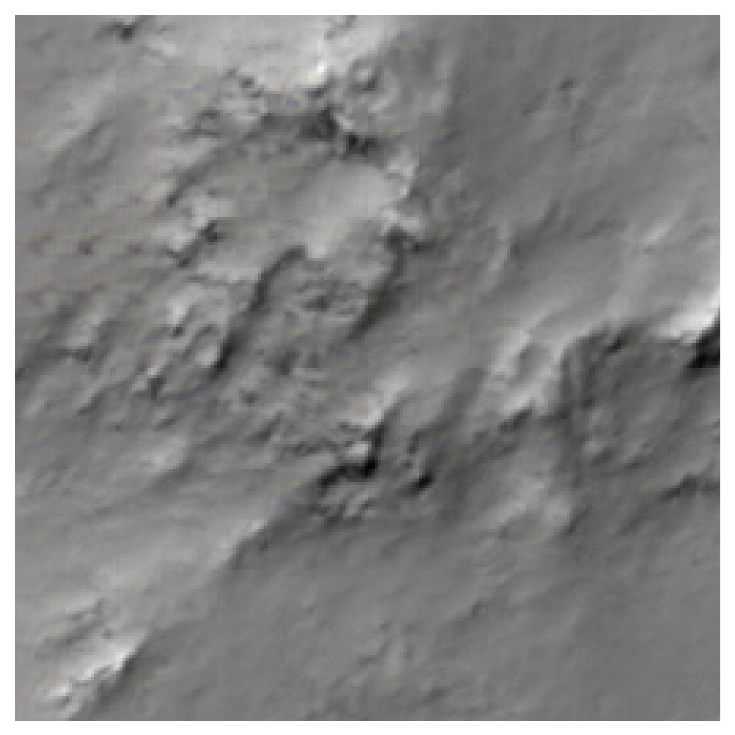}%
\includegraphics[width=0.15\textwidth]{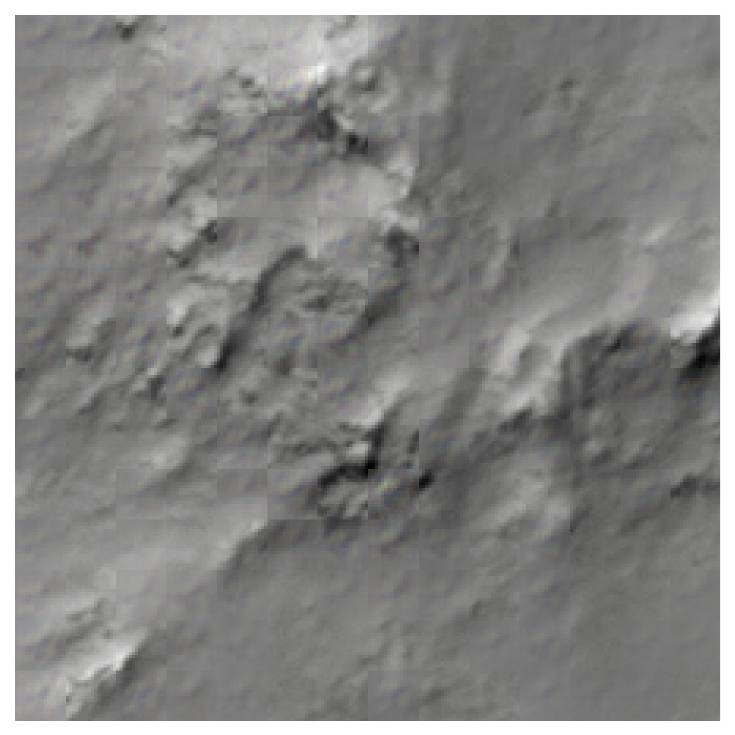}%
\includegraphics[width=0.15\textwidth]{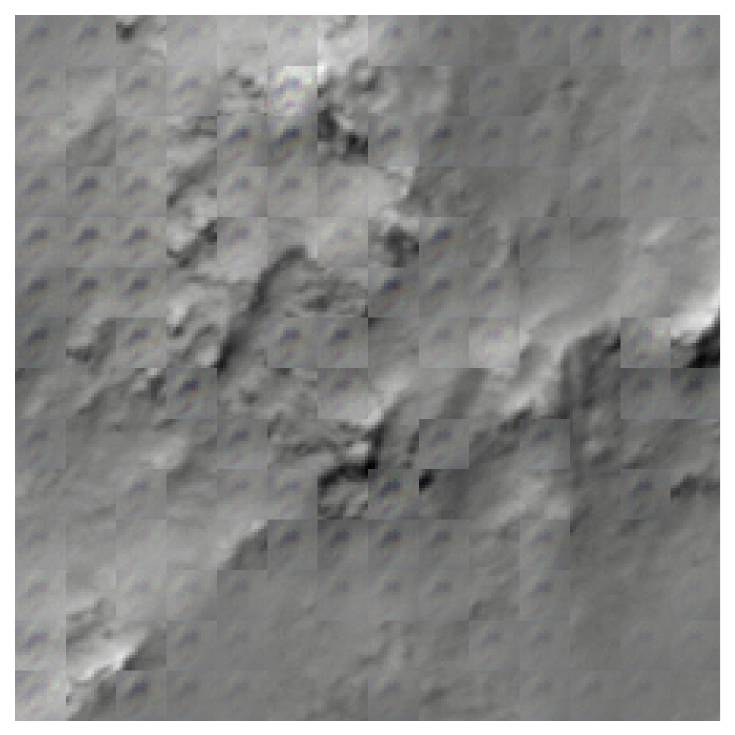}\\
[3pt]
\includegraphics[width=0.15\textwidth]
{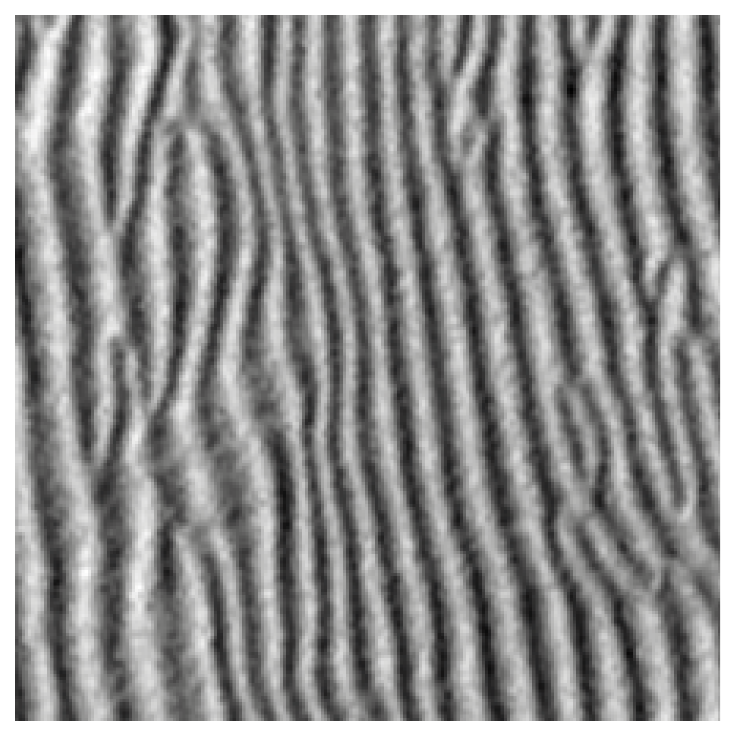}%
\includegraphics[width=0.15\textwidth]{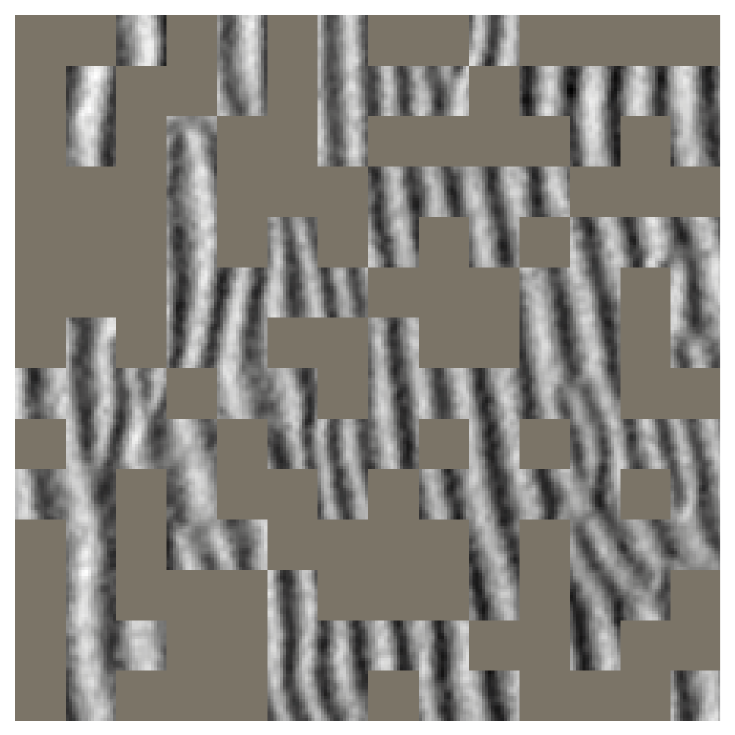}%
\includegraphics[width=0.15\textwidth]{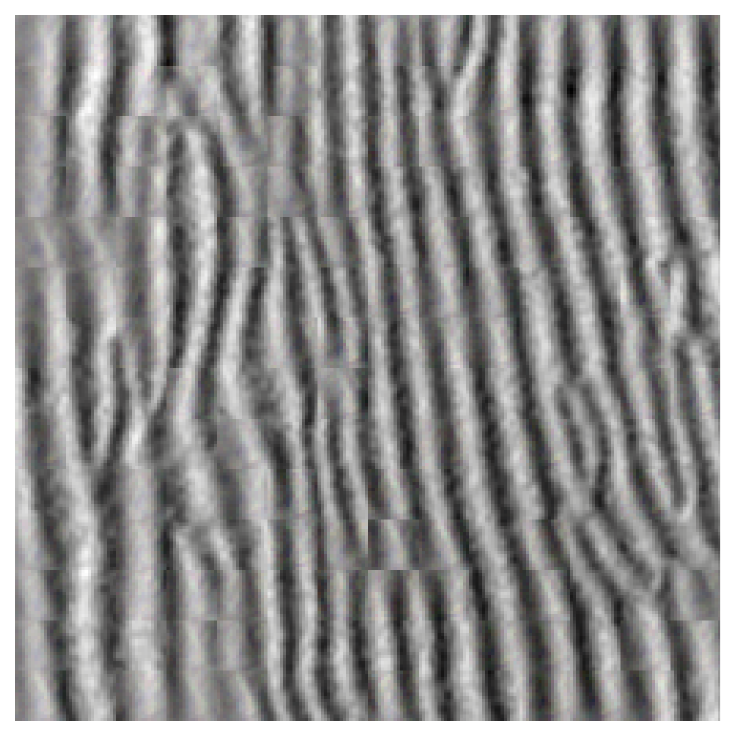}%
\includegraphics[width=0.15\textwidth]{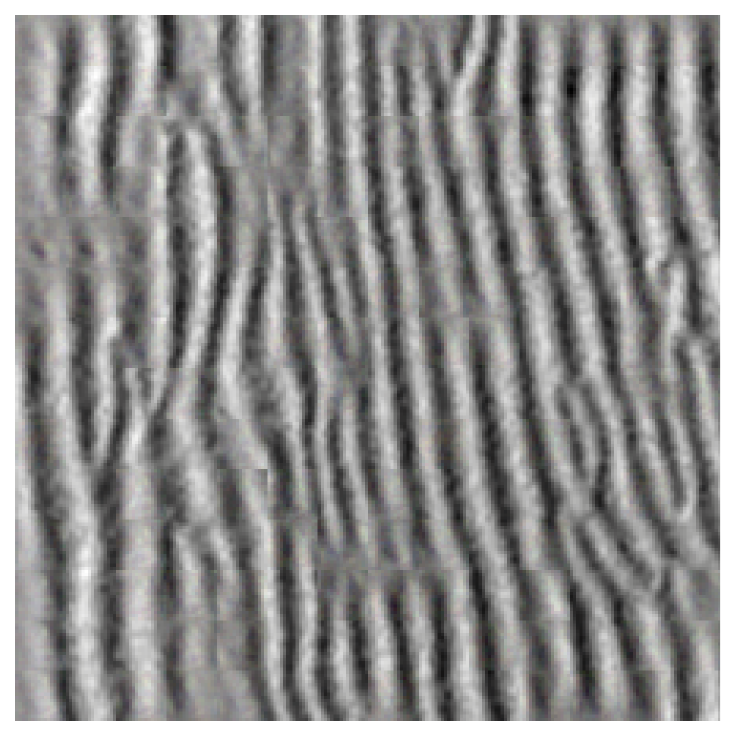}%
\includegraphics[width=0.15\textwidth]{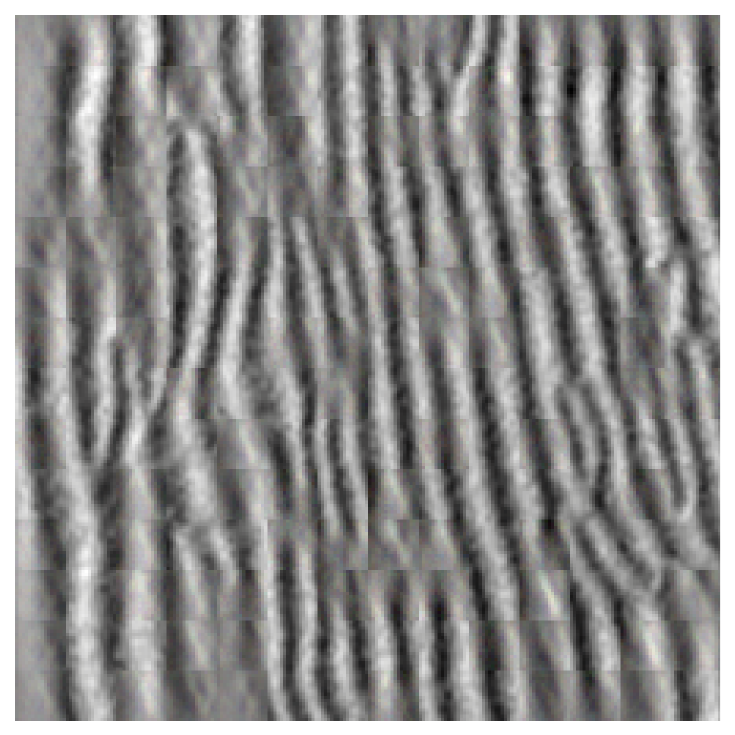}%
\includegraphics[width=0.15\textwidth]{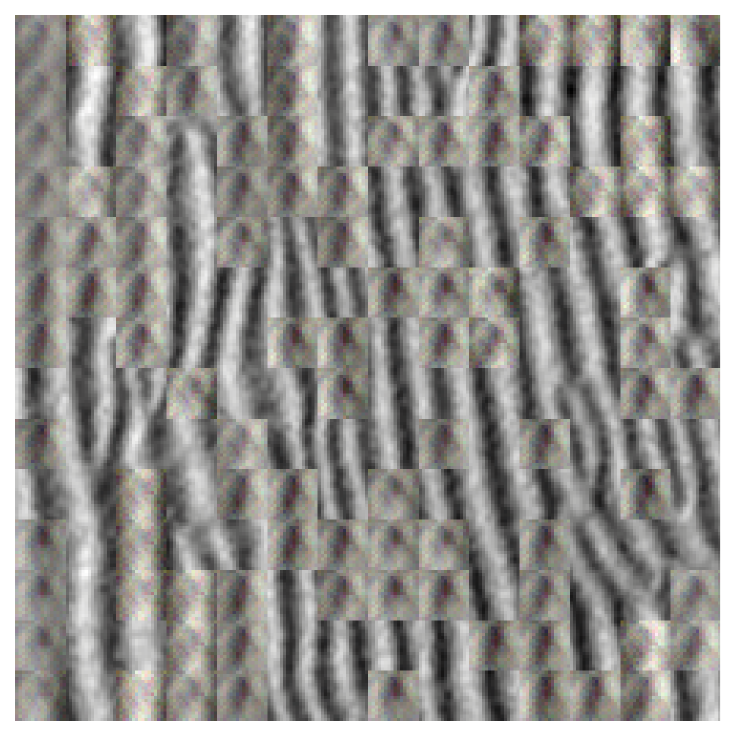}
\caption*{\textbf{HiRISE data sample reconstruction}}
\end{subfigure}
\hfill
\begin{subfigure}{0.9\textwidth}
\centering
\includegraphics[width=0.15\textwidth]{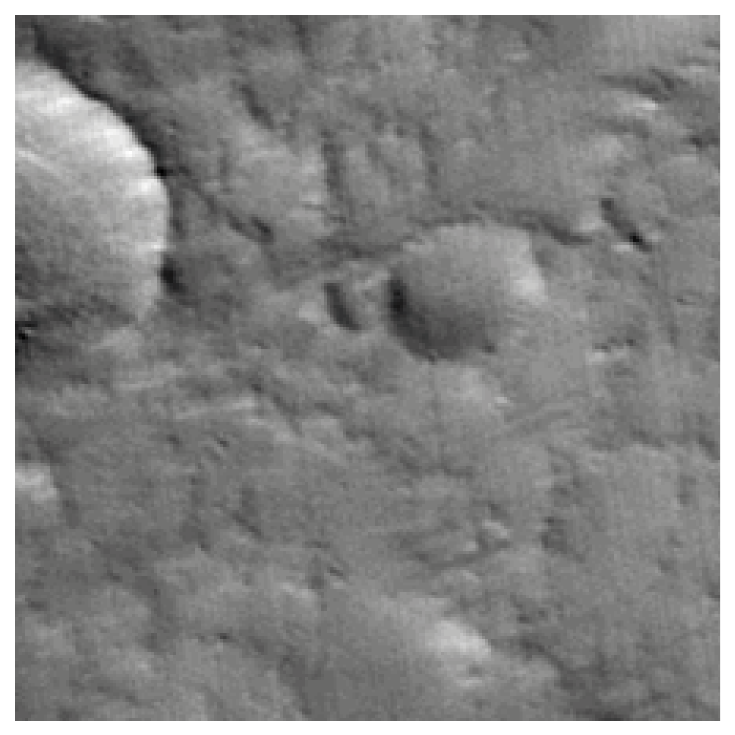}%
\includegraphics[width=0.15\textwidth]{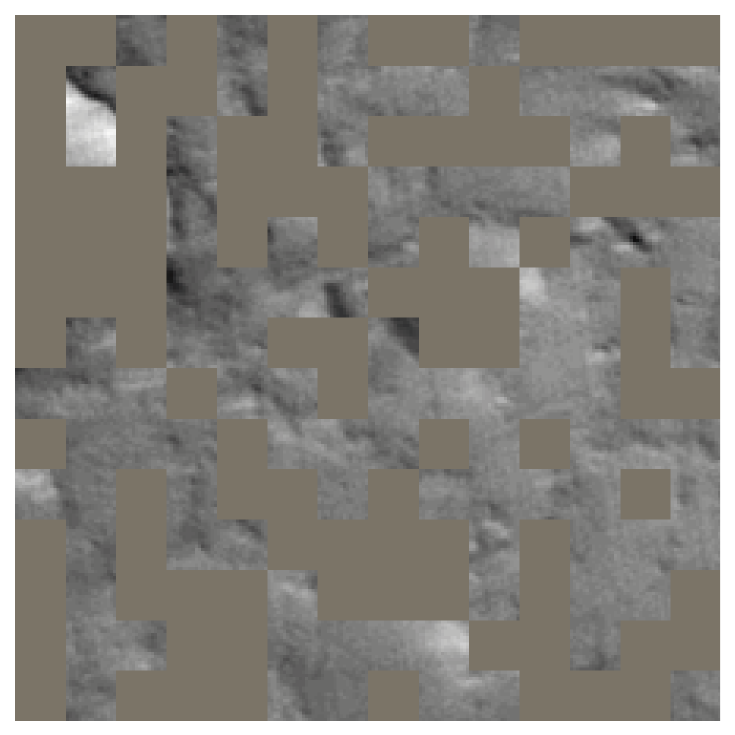}%
\includegraphics[width=0.15\textwidth]{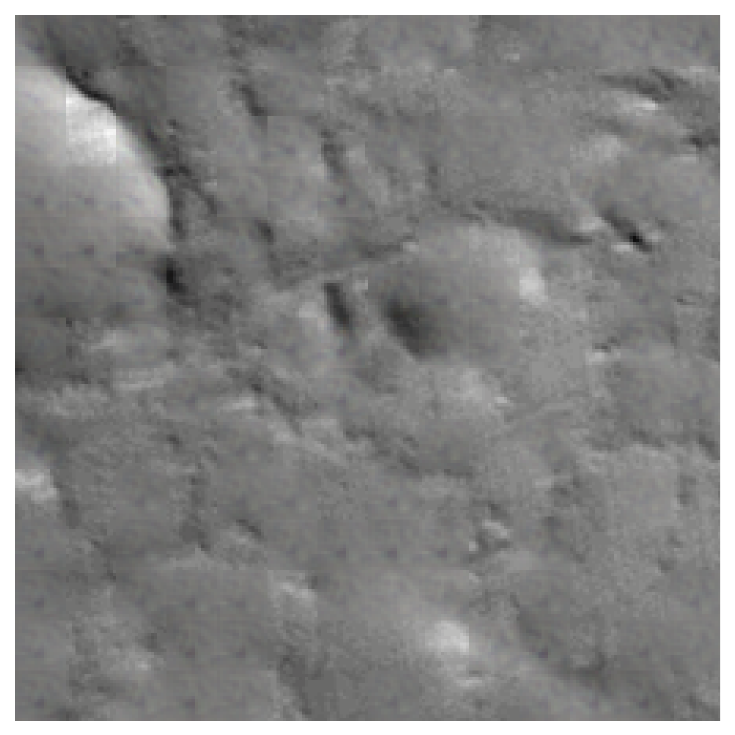}%
\includegraphics[width=0.15\textwidth]{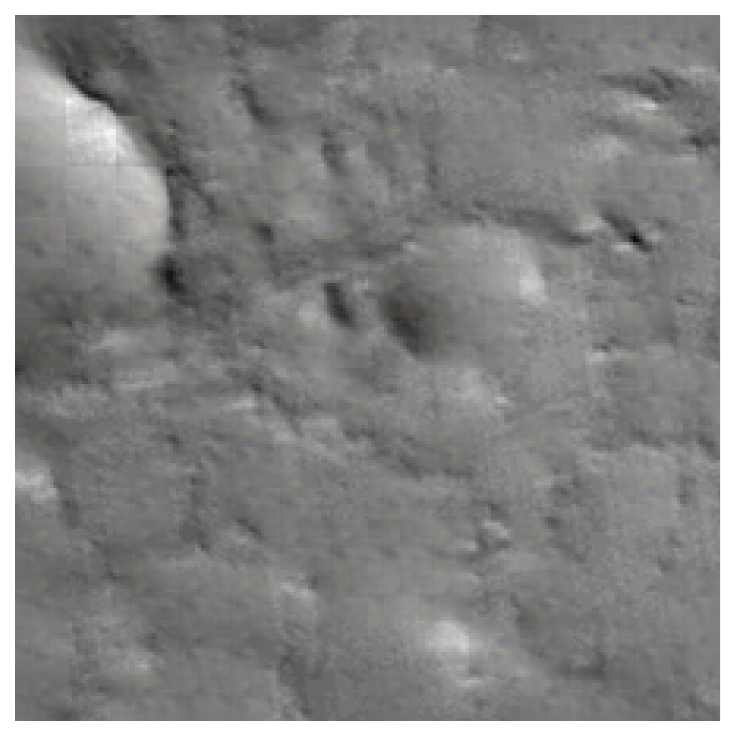}%
\includegraphics[width=0.15\textwidth]{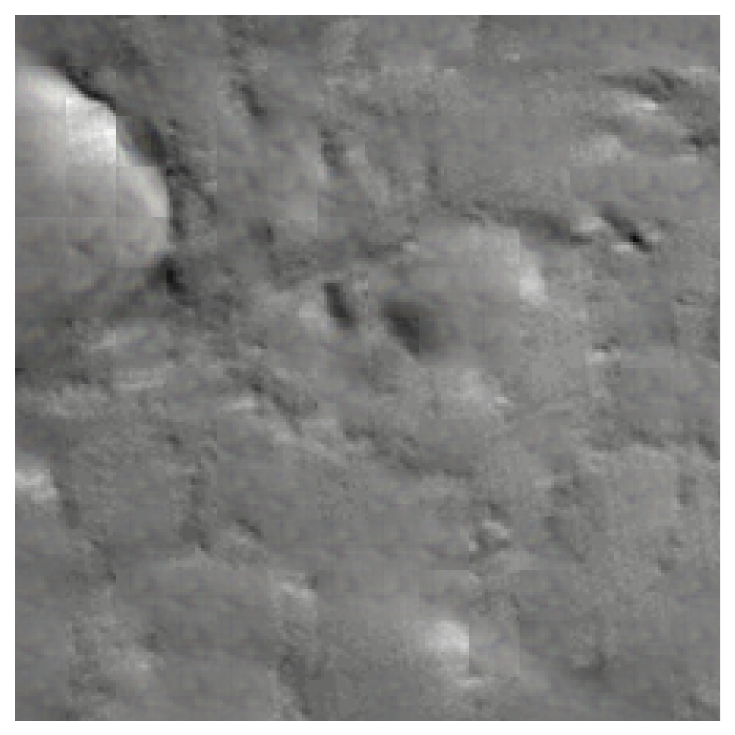}%
\includegraphics[width=0.15\textwidth]{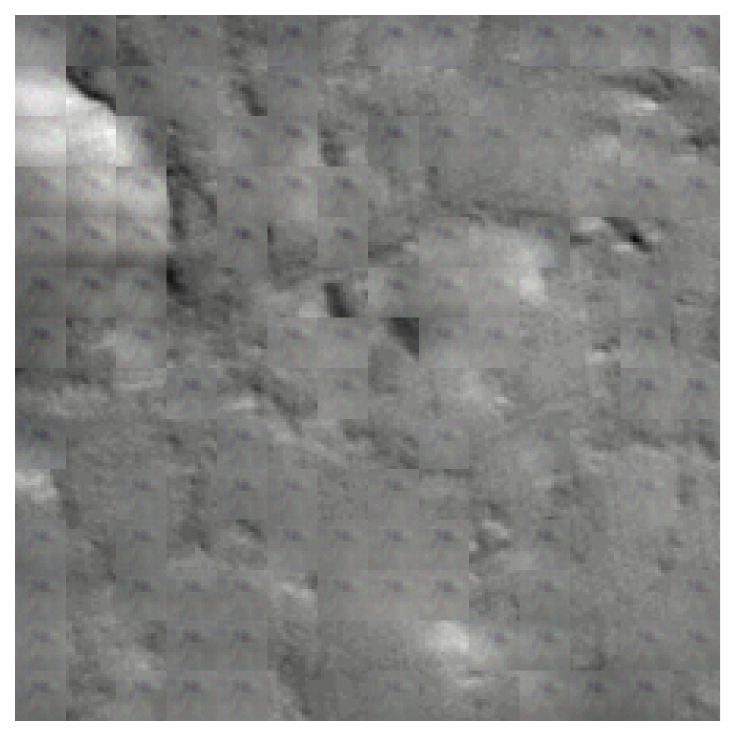}\\
[3pt]
\includegraphics[width=0.15\textwidth]{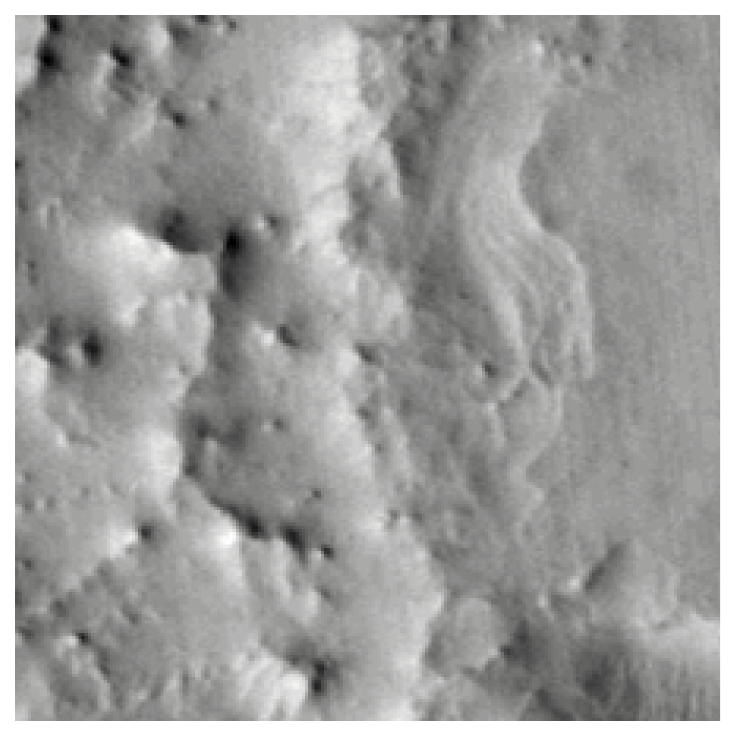}%
\includegraphics[width=0.15\textwidth]{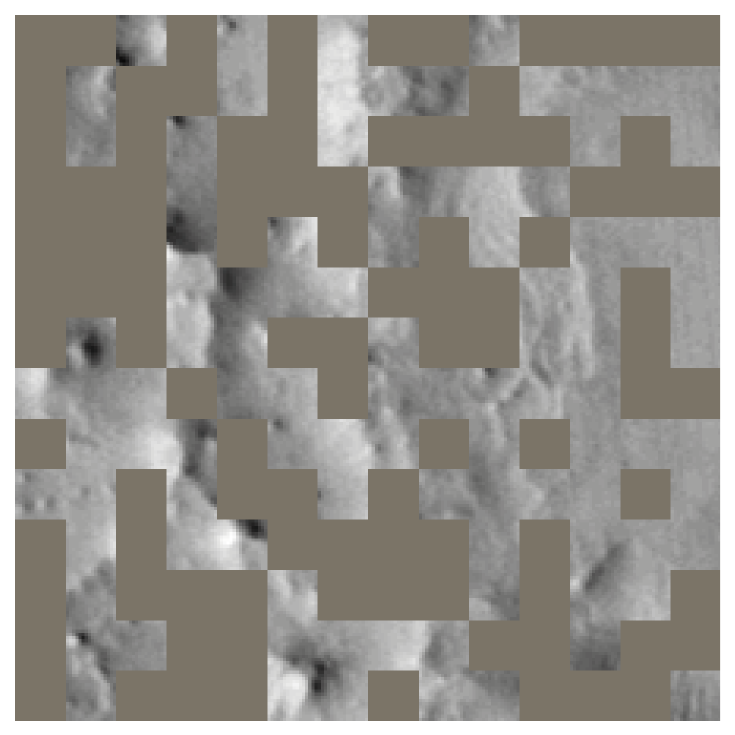}%
\includegraphics[width=0.15\textwidth]{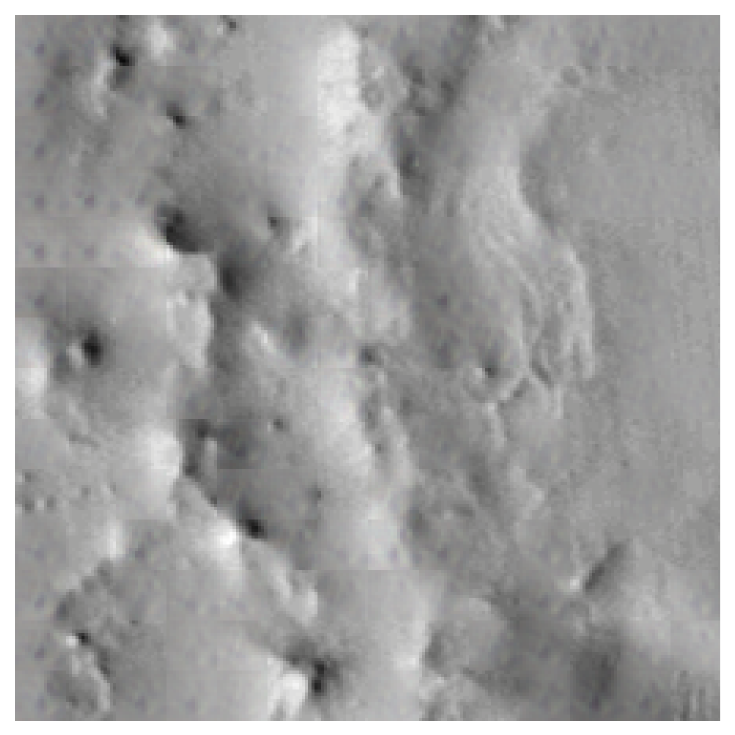}%
\includegraphics[width=0.15\textwidth]{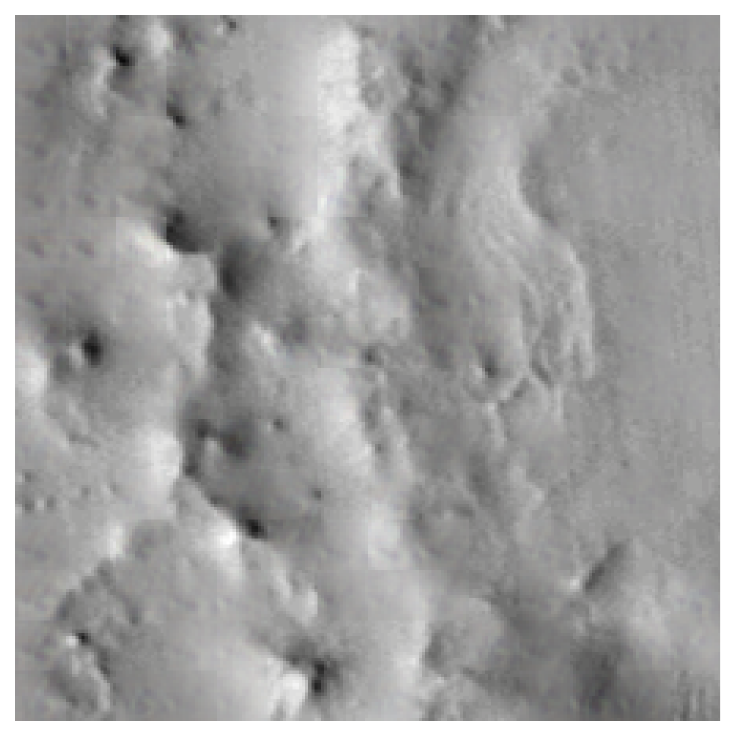}%
\includegraphics[width=0.15\textwidth]{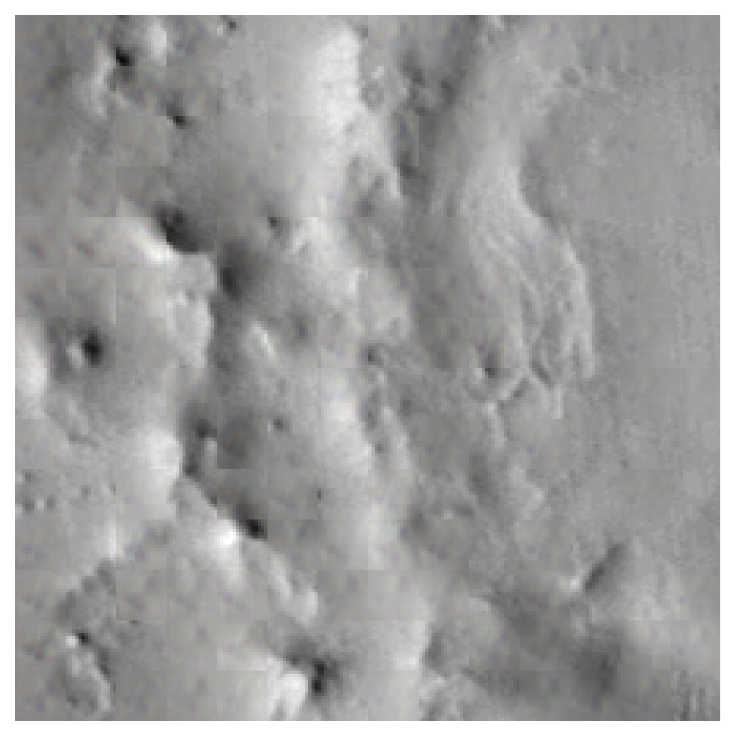}%
\includegraphics[width=0.15\textwidth]{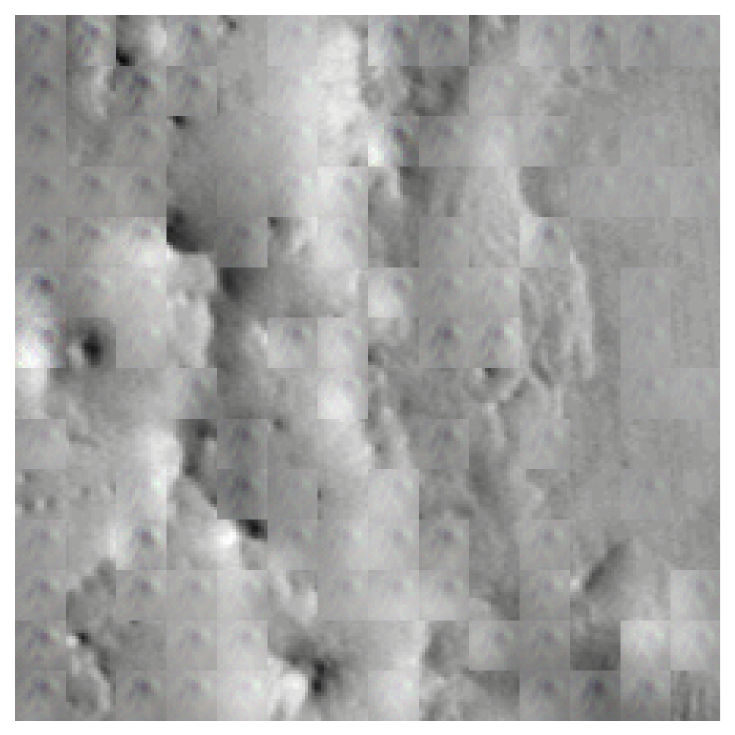}
\caption*{\textbf{CTX data sample reconstruction}}
\end{subfigure}

\begin{subfigure}{0.9\textwidth}
\centering
\includegraphics[width=0.15\textwidth]{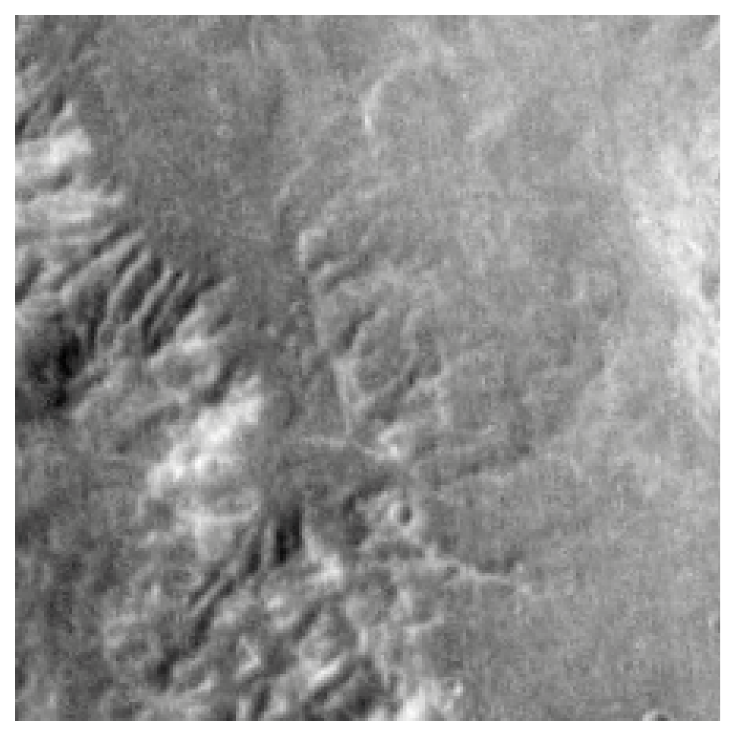}%
\includegraphics[width=0.15\textwidth]{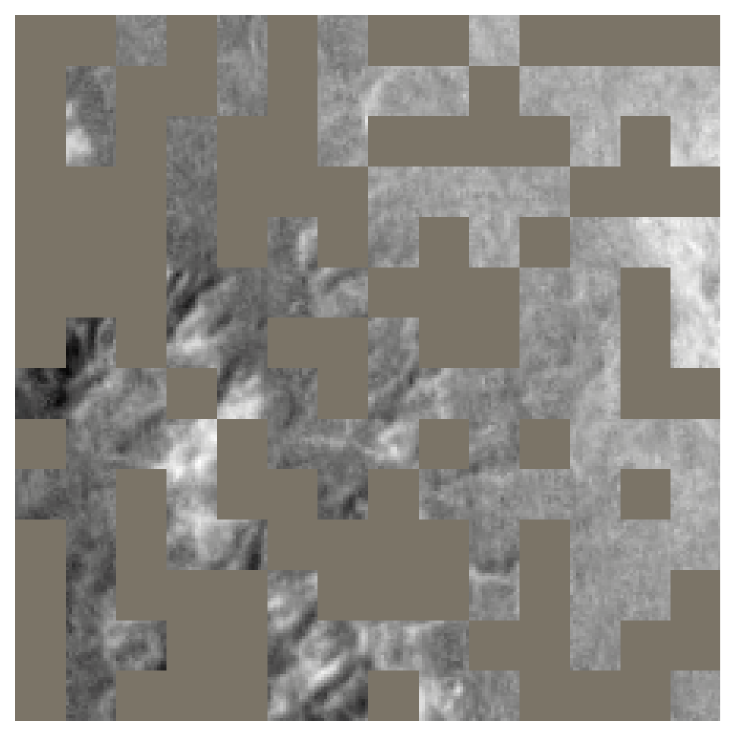}%
\includegraphics[width=0.15\textwidth]{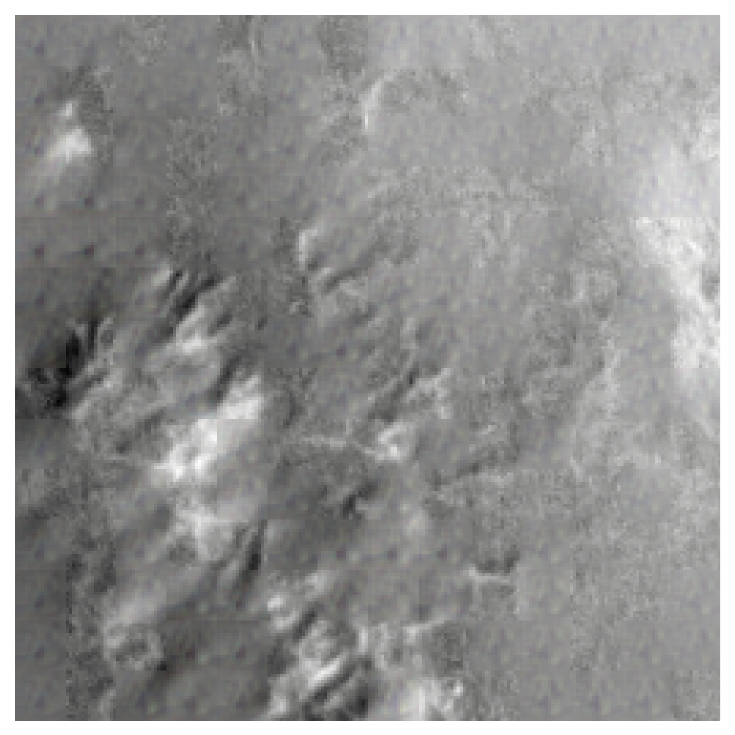}%
\includegraphics[width=0.15\textwidth]{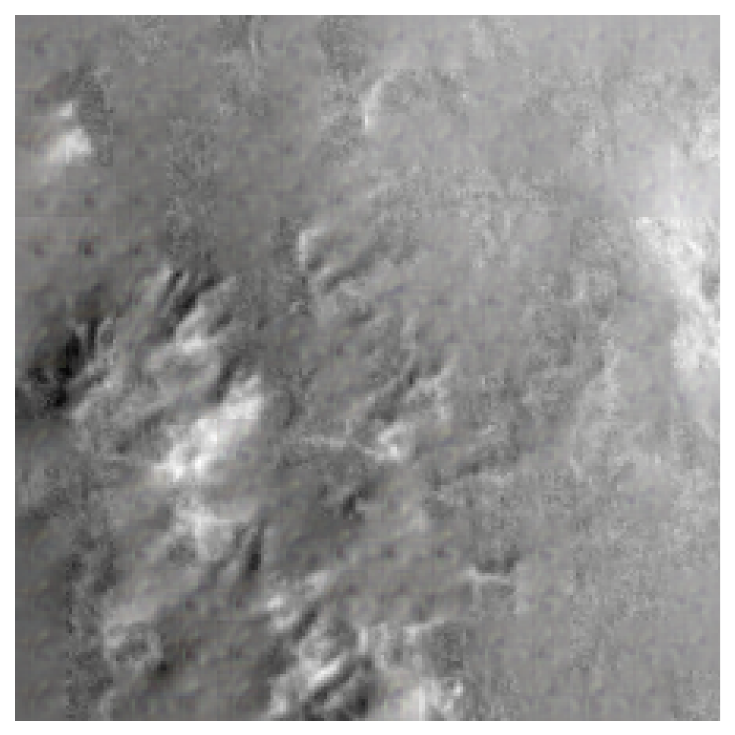}%
\includegraphics[width=0.15\textwidth]{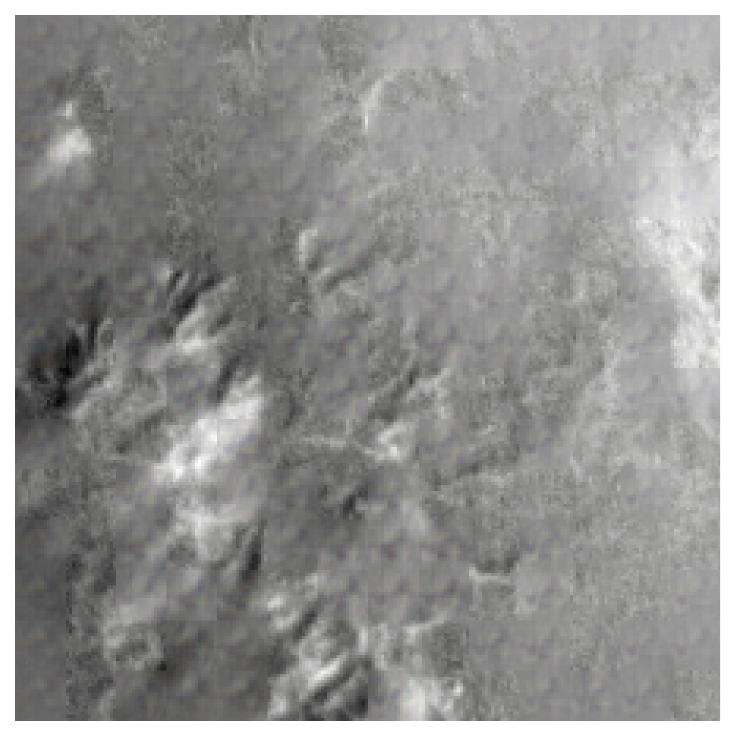}%
\includegraphics[width=0.15\textwidth]{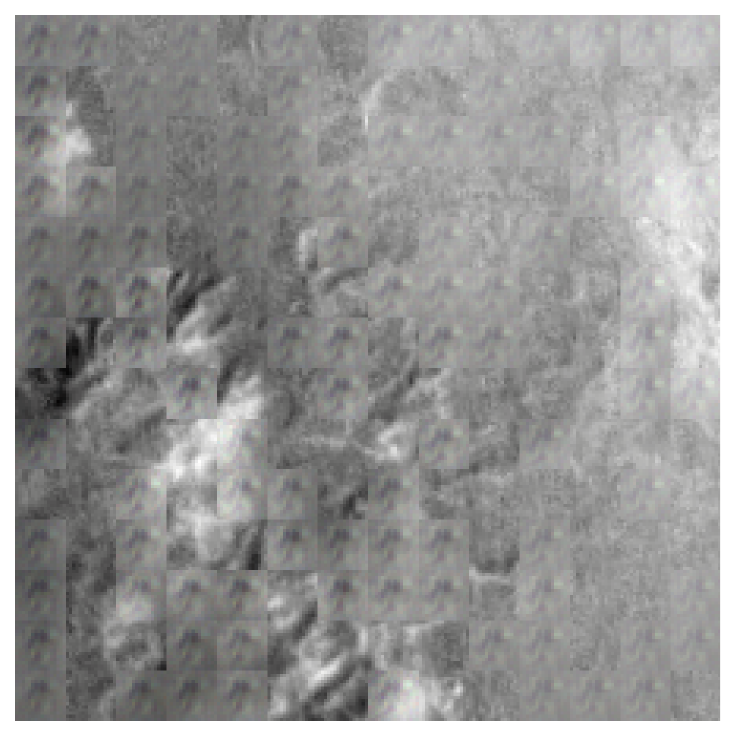}\\
[3pt]
\includegraphics[width=0.15\textwidth]{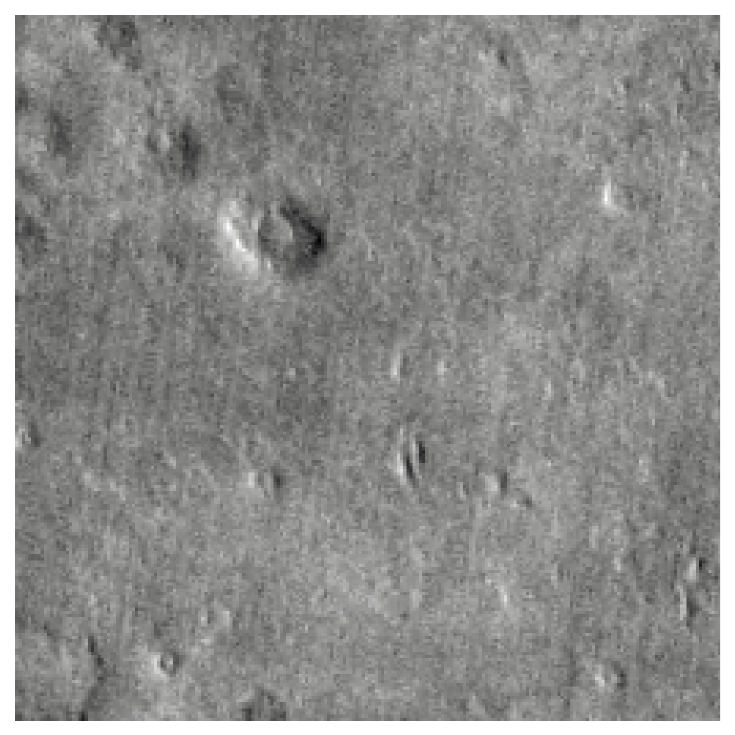}%
\includegraphics[width=0.15\textwidth]{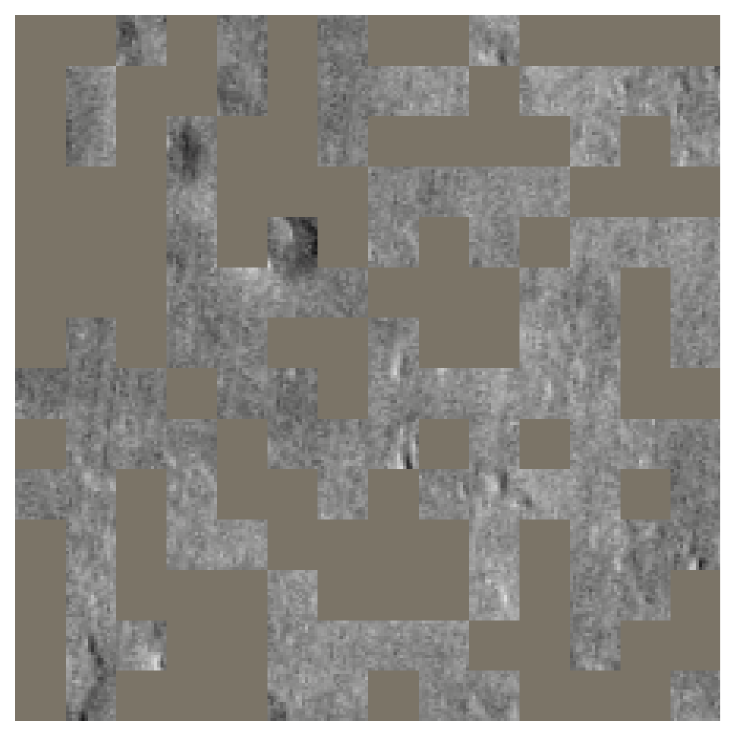}%
\includegraphics[width=0.15\textwidth]{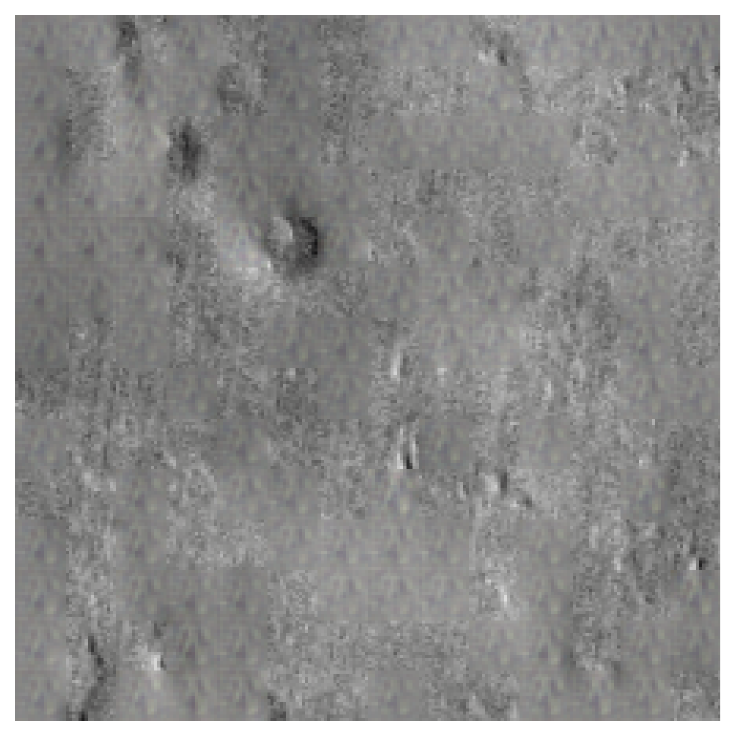}%
\includegraphics[width=0.15\textwidth]{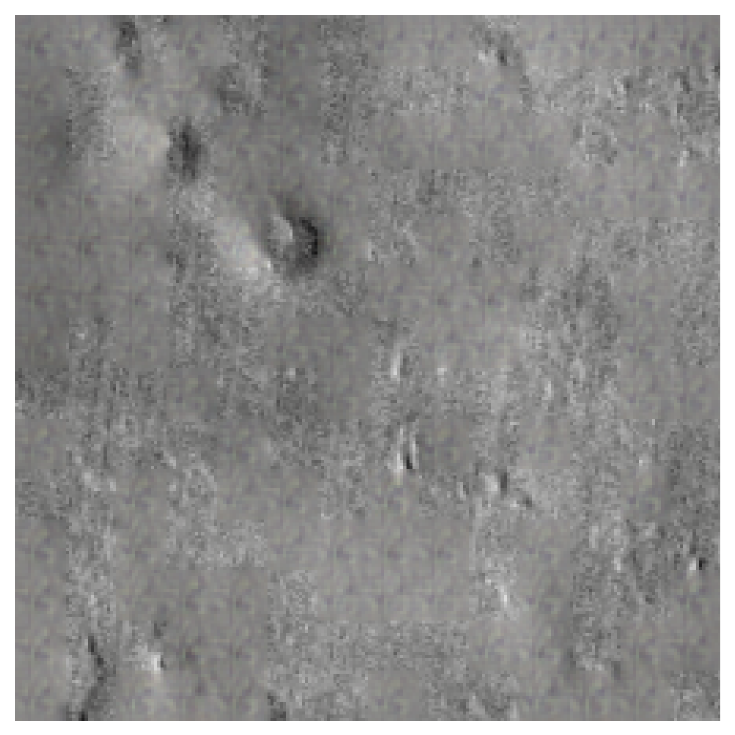}%
\includegraphics[width=0.15\textwidth]{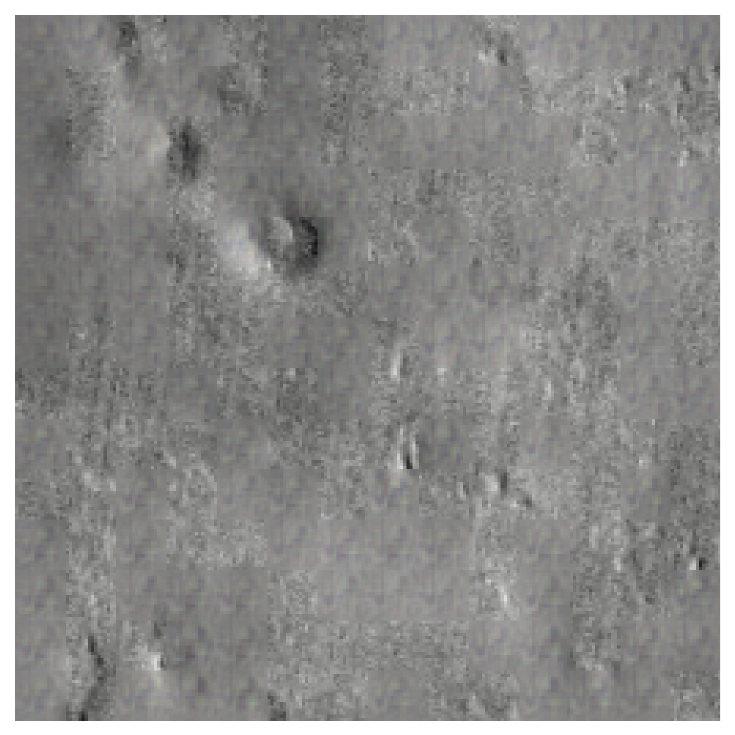}%
\includegraphics[width=0.15\textwidth]{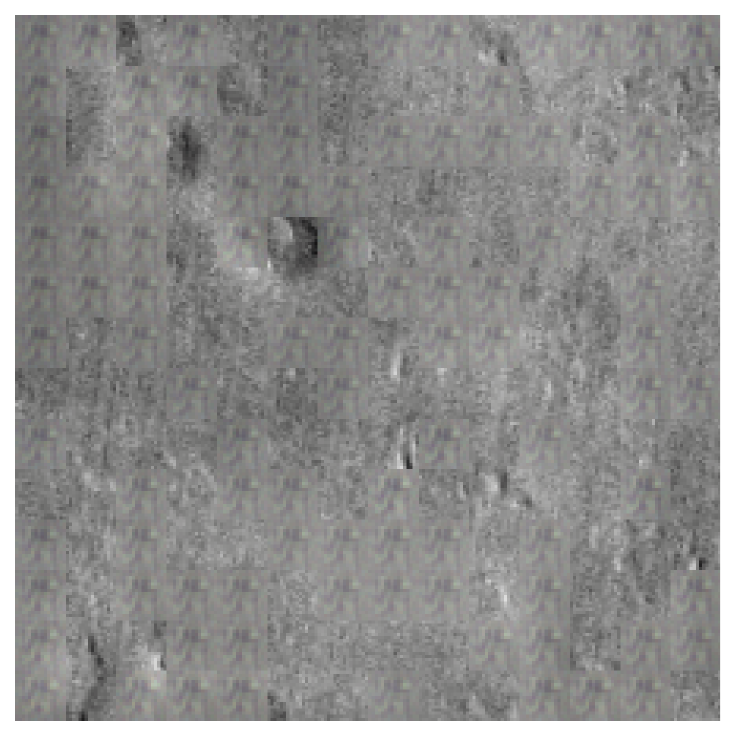}
\caption*{\textbf{THEMIS data sample reconstruction}}
\end{subfigure}

\caption{Reconstruction results using models pre-trained with the proposed combined loss function (pixel-based + perceptual). This figure visualizes reconstructions for data samples from all three sensors: HiRISE (top rows), CTX (middle rows), and THEMIS (bottom rows). The columns display the Original image, the Masked input, and the outputs from the individual sensor models and the HCT model.}
\label{fig:my_loss_reconstruction}
\end{figure*}

In contrast, Figure \ref{fig:my_loss_reconstruction} shows reconstructions from models pre-trained using our proposed combined loss. We visualize two samples from each of the three sensors. Across all sensors, the reconstructions capture not only the correct color distribution but also the underlying surface morphology with substantially higher clarity. These results highlight the effectiveness of our loss formulation in guiding the model to learn feature-aware representations that preserve critical geomorphological structures.

\newpage

\subsection{Scaling coefficient}
\label{subsec:scaling_coefficient}

To analyze the sensitivity of our method to the scaling coefficient used during model merging, we conducted experiments by varying the coefficient from 0.1 to 1.0 in increments of 0.1. These experiments were performed only on downstream tasks that showed significant differences compared to baselines and among different checkpoint selection strategies. Hence, binary classification datasets and \textit{Boulder} and \textit{ConeQuest} segmentation tasks were excluded.

\begin{wrapfigure}{r}{0.45\textwidth}
    \centering
    \includegraphics[width=0.45\textwidth]{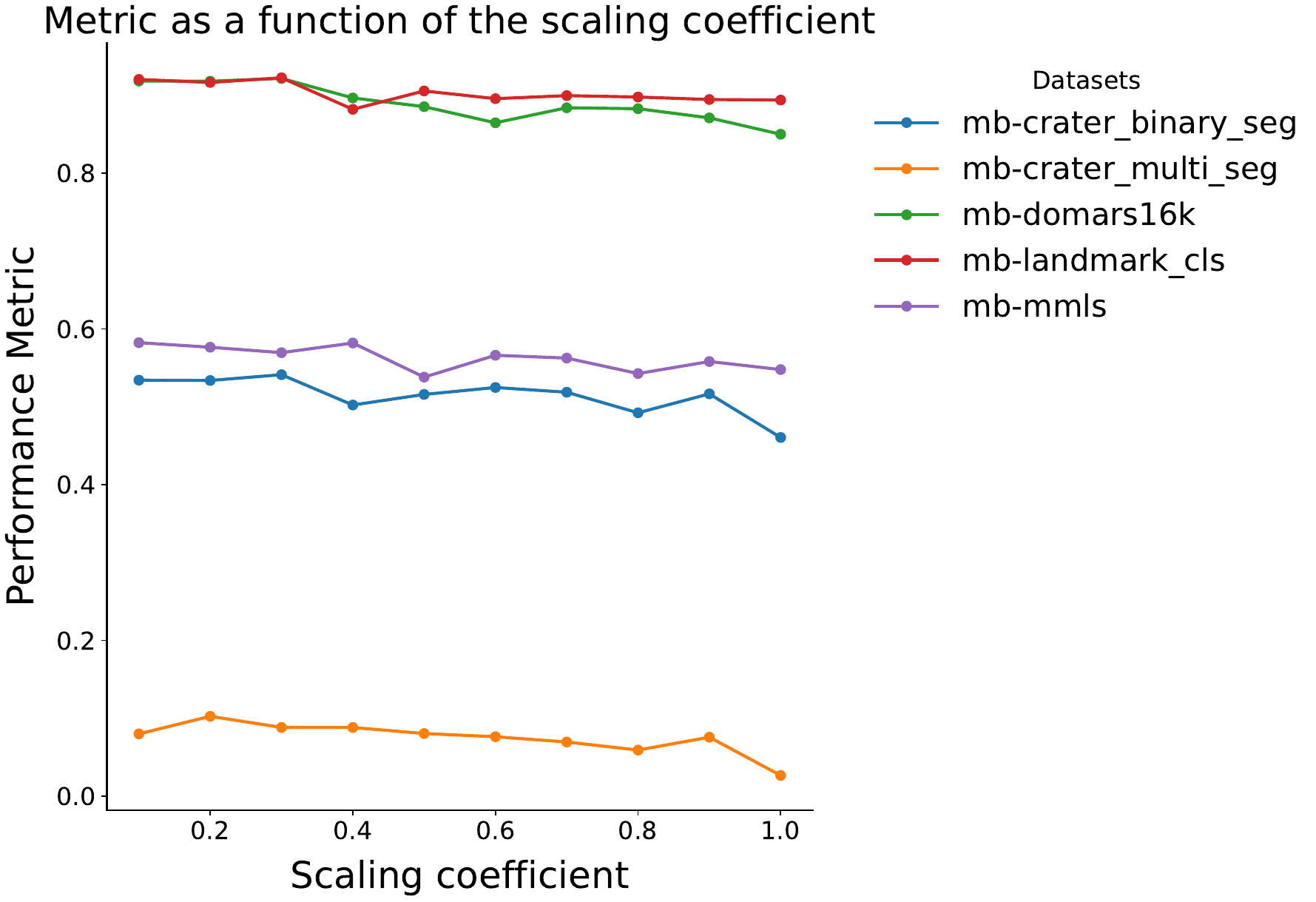}
    \caption{Performance as a function of the scaling coefficient on classification and segmentation downstream tasks.}
    \label{fig:scaling_coefficient}
\end{wrapfigure}

Figure \ref{fig:scaling_coefficient} presents the results for both classification and segmentation tasks, where we report the F1-Score for classification and mIoU for segmentation. As shown in the figure, the performance of the proposed approach remains largely stable across different scaling coefficients, indicating that our method is not highly sensitive to this parameter. This observation is consistent with the findings reported by \citet{ilharco2022editing}. Additionally, as the scaling coefficient increases beyond a certain threshold, performance decreases across most datasets, indicating that excessively high scaling values are not beneficial, again consistent with \citet{ilharco2022editing}.

\FloatBarrier
\subsection{Ablation on Tolerance hyperparameter ($\epsilon$)}
\label{subsec:epsilon}

To evaluate the sensitivity of our method to the tolerance hyperparameter ($\epsilon$), we conduct experiments by varying its value to $10^{-2}$ and $10^{-3}$, and compare these results with the default setting of $10^{-4}$. The results are reported in Table \ref{tab:epsilon}. We observe that changing $\epsilon$ has minimal impact on performance across most datasets, with results either remaining consistent or improving slightly by 1–2\%. The only exception is the \textit{ConeQuest} dataset, where performance decreases marginally; however, the drop is limited to approximately 2\%, indicating that the method remains robust to variations in $\epsilon$.






\begin{table}[htbp]
\centering

\begin{minipage}{0.48\textwidth}
\centering
\footnotesize
\setlength{\tabcolsep}{3pt}
\renewcommand{\arraystretch}{0.9}
\begin{tabular}{lcccc}
\toprule
\textbf{$\epsilon$} & \textbf{DoMars16k} & \textbf{Landmark} & \textbf{ConeQuest} & \textbf{Crater Multi} \\
\midrule
$10^{-2}$ & 0.92 & 0.92 & 0.70 & 0.15 \\
$10^{-3}$ & 0.93 & 0.94 & 0.69 & 0.15 \\
$10^{-4}$ & 0.92 & 0.91 & 0.71 & 0.14 \\
\bottomrule
\end{tabular}
\raggedright
\captionof{table}{Results for different values of the tolerance \\hyperparameter ($\epsilon$).}
\label{tab:epsilon}
\end{minipage}
\hfill
\begin{minipage}{0.49\textwidth}
\centering
\footnotesize
\setlength{\tabcolsep}{3pt}
\renewcommand{\arraystretch}{0.9}
\begin{tabular}{lcccc}
\toprule
 & \textbf{DoMars16k} & \textbf{Landmark} & \textbf{ConeQuest} & \textbf{Crater Multi} \\
\midrule
(H + C) + T & 0.92 & 0.93 & 0.69 & 0.15 \\
MOMO        & 0.92 & 0.91 & 0.71 & 0.14 \\
\bottomrule
\end{tabular}
\captionof{table}{Results for incremental sensor merging, where a THEMIS model is merged with an existing HiRISE and CTX model ((H + C) + T), compared with MOMO.}
\label{tab:new_sensor}
\end{minipage}

\end{table}

\subsection{Merging New Modality}
\label{subsec:add_new_sensor}

To evaluate how performance is affected when incorporating a new sensor, we conduct an experiment simulating incremental sensor addition. In this setup, we assume access to independently trained models along with their validation loss trajectories. We first consider models trained on HiRISE and CTX as existing sensors, and then introduce THEMIS as a new sensor modality. Based on the validation trajectory of the THEMIS model, we select the checkpoint whose validation loss is closest to that of the existing models and merge it accordingly.

Due to computational constraints, we report results on two classification datasets and two segmentation datasets. The results are summarized in Table \ref{tab:new_sensor}. We observe that incorporating the new sensor does not significantly affect performance, with changes remaining within $\pm 1$-$2\%$ across all evaluated tasks.

\subsection{Research Impact}
\label{subsec:research_impact}

In this section, we discuss real-world use cases of \textbf{MOMO}.

\subsubsection{Comparison with PDS deployed Model}
\label{subsubsec:pds_comparison}

The NASA Planetary Data System (PDS) archives data from planetary science missions, and its Cartography and Imaging Sciences Node (Imaging Node) provides public access to millions of planetary images. To help scientists search for images based on visual content rather than metadata alone, the Imaging Node introduced a content-based image search capability in 2017. This system, developed using machine learning classification techniques by \citet{wagstaff2021mars}, enables researchers to efficiently identify images relevant to their investigations.

\begin{table*}[htbp]
\resizebox{\textwidth}{!}{
\begin{tabular}{l|cccccccc|c}
\toprule[1.5pt]
& \multicolumn{1}{c}{\textbf{Bright dune}} &
  \multicolumn{1}{c}{\textbf{Crater}} &
  \multicolumn{1}{c}{\textbf{Dark dune}} &
  \multicolumn{1}{c}{\textbf{Impact ejecta}} &
  \multicolumn{1}{c}{\textbf{Other}} &
  \multicolumn{1}{c}{\textbf{Slope Streak}} &
  \multicolumn{1}{c}{\textbf{Spider}} &
  \multicolumn{1}{c}{\textbf{Swiss cheese}} &
  \multicolumn{1}{c}{\textbf{Macro Avg}} \\

\midrule[1pt]
PDS  & 0.86 & \textbf{0.79} & 0.87 & 0.30 & \textbf{0.96} & 0.67 & 0.04 & 0.94 & 0.68 \\
\rowcolor{lighttan}
MOMO & \textbf{0.90} & 0.75 & \textbf{0.91} & \textbf{0.40} & \textbf{0.96} & \textbf{0.78} & \textbf{0.05} & \textbf{0.99} & \textbf{0.72} \\
\bottomrule[1pt]
\end{tabular}
}
\caption{Per-class F1-scores for PDS and MOMO models on the PDS dataset. \textbf{Bold} numbers indicate the higher F1-score for each class.}
\label{tab:pds_results}
\end{table*}

We compare \textbf{MOMO} with the model currently deployed at NASA’s Planetary Data System (PDS) \cite{wagstaff2021mars}, focusing on the landmark classification dataset used by the PDS Imaging Node. As shown in Table \ref{tab:pds_results}, \textbf{MOMO} outperforms the PDS model across most classes, achieving higher F1-scores in seven out of eight categories and improving the overall macro-average by 4\%. Notably, MOMO shows significant improvements in \textit{Slope Streak}, \textit{Impact ejecta}, and \textit{Swiss cheese}, with gains of 11\%, 10\%, and 5\%, respectively, demonstrating its effectiveness in capturing complex surface morphologies and fine-grained Martian features. Although the PDS model performs slightly better on the \textit{Crater} class, MOMO achieves more balanced and consistent performance across diverse geologic feature types, making it a stronger candidate for large-scale automated mapping and planetary data analysis.

\subsubsection{Creating Global Maps}
\label{subsubsec:global_maps}

Scientists and planetary geologists are interested in studying geologic features on Mars and understanding their global distribution. To achieve this, they typically create small labeled datasets and train machine learning models to generate global maps of specific features. Given its strong segmentation performance, \textbf{MOMO} can serve as an effective tool for producing such large-scale global maps of Martian surface features.

\begin{wrapfigure}{r}{0.58\textwidth}
    \centering
    \includegraphics[width=0.58\textwidth]{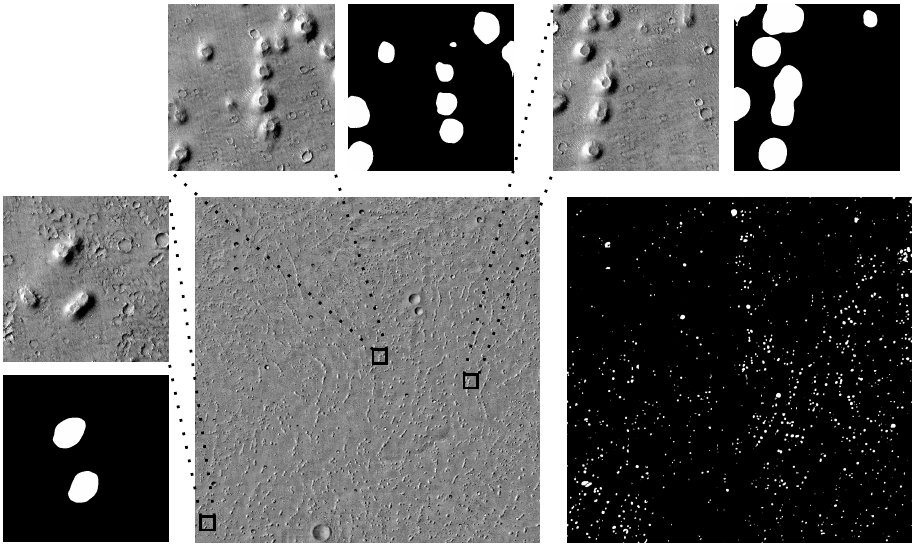}
    \caption{Example of global map generation using \textbf{MOMO} on the out-of-distribution region of the \textit{ConeQuest} dataset. The center panel shows the original large-scale HiRISE tile, and the right panel shows the stitched prediction map after inference. The left and top panels display representative 512×512 data samples and their corresponding segmentation outputs. This experiment demonstrates MOMO’s capability to generalize to unseen regions and its potential for large-scale planetary surface mapping.}
    \label{fig:global_map}
\end{wrapfigure}

To demonstrate the efficiency and practical utility of \textbf{MOMO}, we perform inference on the \textit{ConeQuest} dataset using out-of-distribution (OOD) data. To replicate this process, we exported new data from JMARS \cite{Christensen2009JMARS}. JMARS (Java Mission-planning and Analysis for Remote Sensing) is a geospatial information system developed to visualize, analyze, and export planetary data from multiple Mars missions, focusing on regions not included in the original training set.

Each data tile in \textit{ConeQuest} provides latitude and longitude information, which allowed us to select a previously unseen region centered at 15° latitude and 84° longitude. We exported CTX imagery covering an area of approximately 1.5 km × 1.5 km (12288 x 12288 pixels), sampled into 512 × 512 pixel tiles with an overlap of 256 pixels, resulting in a total of 2,306 image samples.

Figure \ref{fig:global_map} illustrates an example of this experiment. The left panel shows the original large-scale tile, and the right panel shows the stitched output generated after performing inference with MOMO. For reference, we also display a few example 512 × 512 tiles used for prediction. These results demonstrate that MOMO can be effectively used to produce global-scale maps of geologic features from unseen regions, highlighting its potential for planetary-scale mapping applications.

\end{document}